\documentclass{article}

\usepackage{arxiv}
\usepackage{xcolor}
\usepackage[utf8]{inputenc} 
\usepackage[T1]{fontenc}    
\usepackage{hyperref}       
\usepackage{url}            
\usepackage{booktabs}       
\usepackage{amsfonts}       
\usepackage{nicefrac}       
\usepackage{microtype}      
\usepackage{lipsum}		
\usepackage{graphicx}
\usepackage{natbib}
\usepackage{doi}
\usepackage{graphicx}
\usepackage{subcaption}
\usepackage{multicol}
\usepackage{multirow}
\usepackage{lipsum}
\usepackage{mwe}
\usepackage{tabularx}

\title{Robustness of Segment Anything Model: A Comprehensive Study on Occlusion, Style and Corruptions}

\title{Robustness of Segment Anything Model: A Comprehensive Study on Occlusion, Style and Corruptions}

\title{Understanding Segment Anything Model: SAM is Biased Towards Texture Rather than Shape} 

\author{
Chaoning Zhang\thanks{You are welcome to contact us through chaoningzhang1990@gmail.com} \\
	Kyung Hee University \\
 \And
Yu Qiao \\
	Kyung Hee University \\
 \And
Shehbaz Tariq \\
	Kyung Hee University \\
 \And
Sheng Zheng \\
	Beijing Institute of Technology \\
 \And
Chenshuang Zhang \\
	KAIST \\
 \And
Chenghao Li \\
	KAIST\\
  \And
Hyundong Shin \\
	Kyung Hee University \\
 \And
Choong Seon Hong\\
Kyung Hee University\\
}

\renewcommand{\headeright}{}
\renewcommand{\undertitle}{}
\renewcommand{\shorttitle}{Understanding Segment Anything Model}

\begin{document}
\maketitle

\begin{abstract} 
In contrast to the human vision that mainly depends on the shape for recognizing the objects, deep image recognition models are widely known to be biased toward texture. Recently, Meta research team has released the first foundation model for image segmentation, termed segment anything model (SAM), which has attracted significant attention. In this work, we understand SAM from the perspective of texture \textit{v.s.} shape. Different from label-oriented recognition tasks, the SAM is trained to predict a mask for covering the object shape based on a promt. With this said, it seems self-evident that the SAM is biased towards shape. In this work, however, we reveal an interesting finding: the SAM is strongly biased towards texture-like dense features rather than shape. This intriguing finding is supported by a novel setup where we disentangle texture and shape cues and design texture-shape cue conflict for mask prediction.
\end{abstract}

\section{Introduction}

Image recognition has made significant strides in the field of computer vision due to the development of deep learning~\cite{lecun2015deep}. With image classification on ImageNet~\cite{deng2009imagenet} as an example, AlexNet~\cite{krizhevsky2012imagenet} is the pioneering CNN-based work to outperform non-deep learning methods by a large margin. After that, numerous models have emerged to push the performance frontiers by making the network deeper~\cite{he2016deep,huang2017densely,zhang2020resnet} or relying on a new architecture termed vision transformer (ViT)~\cite{dosovitskiy2020image}. There exist various explanations for how deep models achieve impressive performance on complex image recognition tasks. On the one hand, a line of work has highlighted the importance of spatial arrange (\textit{i.e.} shape) of objects~\cite{kriegeskorte2015deep}. On the other hand, it has been pointed out in some works that local textures are more important for the recognition performance. Specifically, it has been shown that the performance almost has no drop after completely destroying the shape features~\cite{baker2018deep}. The above two explanations are termed shape hypothesis and texture hypothesis, which are contradictory. ~\cite{geirhos2018imagenet} resolves this issue by designing texture-shape cue conflict experiments and show that ImageNet trained CNNs are biased towards texture rather than shape. ViTs~\cite{dosovitskiy2021an} are found to show a similar trend of being biased towards texture~\cite{naseer2021intriguing}. Moreover, the origin of the texture bias phenomenon has been analyzed in~\cite{hermann2020origins} which shows that it is prevalent regardless of the architectures and training objectives (supervised or self-supervised). 

Very recently, Meta research team has released a new type of image recognition model termed ``segment anything model (SAM)~\cite{kirillov2023segment}". Different from prior label-oriented recognition tasks (like image classification, object detection, semantic segmentation), the SAM conducts label-free mask prediction based on a prompt (like point or box). Given that the SAM is optimized to generate a mask that covers the object shape, it is self-evident that object shape plays a critical role in segmenting the mask for the object on the image, which well aligns with the human vision. However, it remains unclear how the texture inside the object boundary might influence the generated mask. In this work, we intend to investigate the influence of shape and texture on the mask prediction performance. To this end, we propose to disentangle shape cue and texture cue as well as design a texture-shape cue conflict for mask prediction. Our results demonstrate that SAM is biased towards texture rather than shape. 

\begin{figure*}[!htbp]
     \centering
         \begin{minipage}[b]{0.18\textwidth}
         \includegraphics[width=\textwidth]{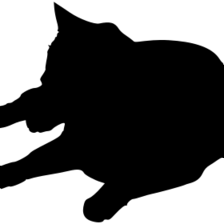}
         \subcaption{cat}
         \end{minipage}
    \begin{minipage}[b]{0.18\textwidth}
         \includegraphics[width=\textwidth, height=1.0\textwidth]       {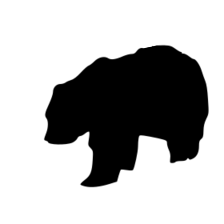}
          \subcaption{bear}
     \end{minipage}
         \begin{minipage}[b]{0.18\textwidth}
         \centering
         \includegraphics[width=\textwidth]{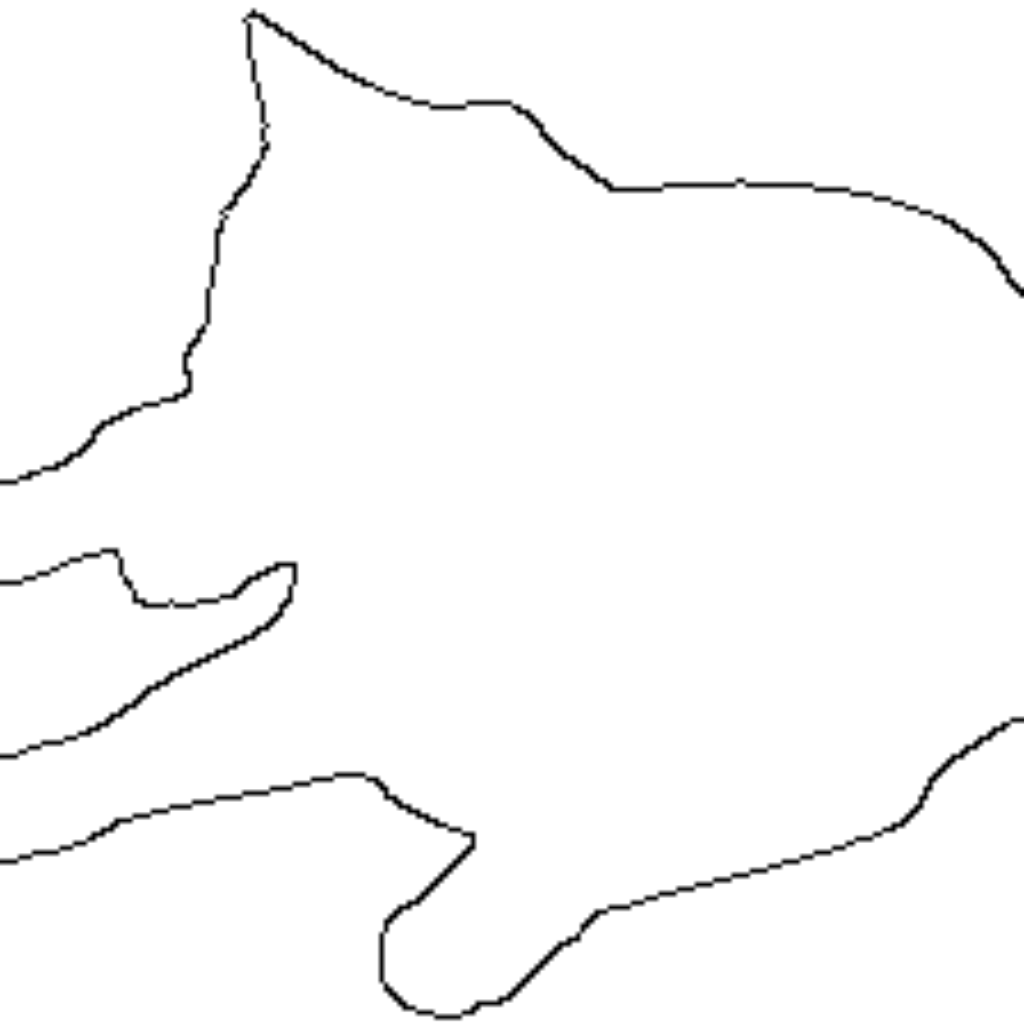}
         \subcaption{shape cue \\ for cat}
     \end{minipage}
        \begin{minipage}[b]{0.18\textwidth}
         \centering
         \includegraphics[width=\textwidth]{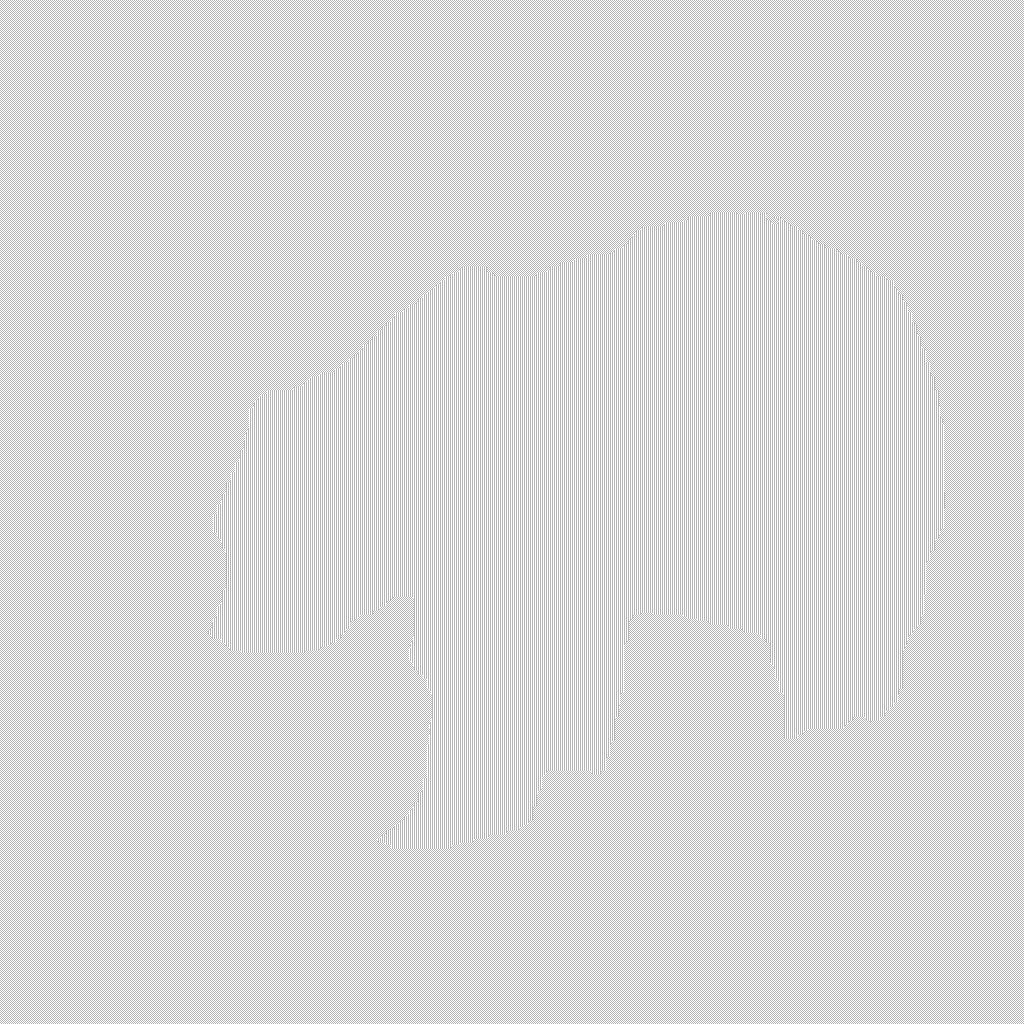}
         \subcaption{texture cue \\ for bear}
     \end{minipage}
    \begin{minipage}[b]{0.18\textwidth}
         \centering
         \includegraphics[width=\textwidth]{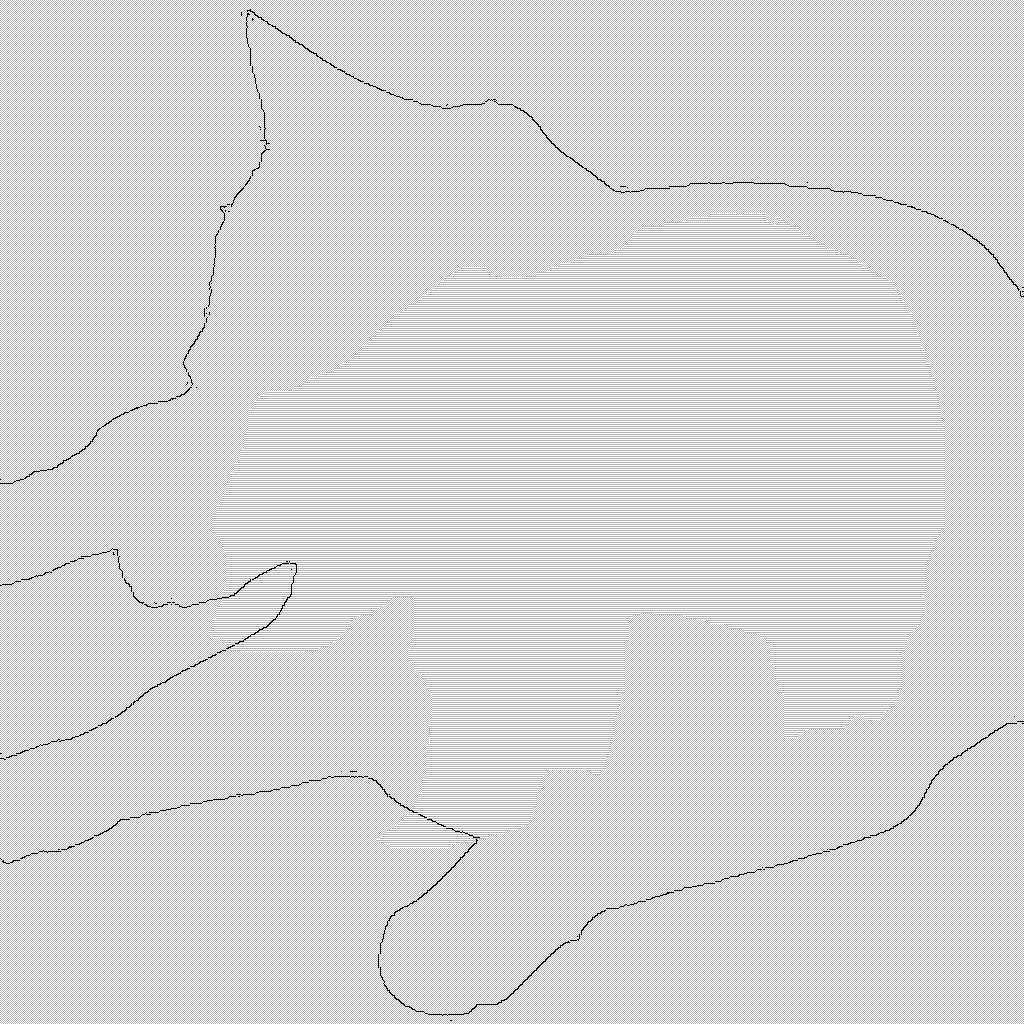}
         \subcaption{texture-shape \\ cue conflict}
     \end{minipage}
        \caption{A concrete example of making texture-shape cue conflict for mask prediction. (a) and (b) represent a cat image and a bear image respectively. (c) represents the shape cue for the cat in (a), while (d) represents texture conflict for (b). (e) combines the shape cue for cat in (c) and texture cue for bear in (d) to finally form a texture-shape cue conflict.}
    \label{fig:conflict_setup}
\end{figure*}

\section{Related works}

\textbf{Segment Anything Model.} Shortly after SAM came out, numerous works have investigated it from different perspectives. Multiple works have investigated the generalization capability of SAM in various challenging scenarios, such as medical images~\cite{ma2023segment,zhang2023input}, Camouflaged Object~\cite{tang2023can} and glass (transparent object and mirror) ~\cite{han2023segment}. SAM is found to often fail in the above scenarios, for which ~\cite{chen2023sam} proposed a task-specific adapter for enhancing its performance. Moreover, a pioneering work~\cite{jing2023segment} investigates preliminary solutions to adapt SAM to Non-Euclidean Domains. Another line of work has focused on improving the utility of SAM. For example, Grounded SAM~\cite{GroundedSegmentAnything2023} realizes the goal of segmenting anything with text inputs via combing Grounding DINO~\cite{liu2023grounding} with SAM. Note that in the original SAM paper, text prompt is also studied but only as a proof-of-concept without providing a public code. Some works~\cite{chen2023semantic,park2023segment} have attempted to combine SAM with BLIP~\cite{li2022blip} or CLIP~\cite{radford2021learning} to assign labels to the generated masks. SAM has also been utilized for image editing~\cite{GroundedSegmentAnything2023,rombach2022high} and inpainting~\cite{yu2023inpaint}, tracking objects in video~\cite{yang2023track,z-x-yang_2023} and reconstructing 3D objects from a single image~\cite{shen2023anything,kang2022any}. Adversarial robustness of SAM has also been investigated by~\cite{zhang2023attacksam}. For a survey on SAM, please refer to~\cite{zhang2023asurvey}.

\textbf{Shape \textit{v.s.} Texture.} Some early works have hypothesized that deep models rely on object shape for image recognition tasks~\cite{kriegeskorte2015deep,lecun2015deep,kubilius2016deep}. This hypothesis is well supported by visualization results that high-level CNN features often encode object parts~\cite{zeiler2014visualizing}. Another early work concluded that deep models resemble children for being biased towards shape rather than color for object classification~\cite{ritter2017cognitive}, for which~\cite{hosseini2018assessing} has presented contrary evidence. In other words, there is another line of works that the model performance of classifying images does not decrease much when the global shape is destroyed not at a cost of performance drop~\cite{gatys2017texture,brendel2019approximating}, but decreases significantly when local texture is removed~\cite{ballester2016performance}. A pioneering study is conducted in~\cite{geirhos2018imagenet} to texture is more dominant than shape for helping recognize image objects. Their finding is supported by a setup which provides texture-shape cue conflict for label prediction. Such a bias towards texture is found to be prevalent in deep recognition models regardless of training objective~\cite{geirhos2018imagenet}. Given that SAM in essence is also a deep recognition model, SAM might also be biased towards texture. However, the task mask prediction highly depends on the object shape, therefore, it seems self-evident that shape plays a dominant role. Therefore, this work intends to resolve these contradictory interpretations by conducting an empirical study.

\section{Background and Method}
\label{sec:occlusion}
Foundation models~\cite{bommasani2021opportunities} have helped push the frontiers of modern AI ranging from NLP to computer vision. NLP has been revolutionalized by BERT~\cite{devlin2018bert} and GPT~\cite{brown2020language,radford2018improving,radford2019language} which are trained on abundant web text. A key difference between BERT~\cite{devlin2018bert} and GPT is that the former requires finetuning on downstream tasks while the latter has strong zero-shot transfer performance. Such text foundation models contribute to the development of various generative AI~\cite{zhang2023complete} tasks including text generation (ChatGPT~\cite{zhang2023ChatGPT} for instance), text-to-image~\cite{zhang2023text} and text-to-speech~\cite{zhang2023audio}, text-to-3D~\cite{li2023generative}. By contrast, the progress of foundation models in the computer vision~\cite{radford2021learning,jia2021scaling,yuan2021florence} lags behind. Two representative breakthroughs are masked autoencoder~\cite{zhang2022survey}, which mimics BERT, and SAM, which follows GPT to adapt the model by prompt. In the following, we briefly summarize how SAM how works.

\textbf{How does SAM work?}
SAM was trained on over 11 million licensed and privacy-preserving images with more than 1 billion masks~\cite{kirillov2023segment}. SAM consists of an image encoder, a prompt encoder and a mask decoder. The image encoder takes an image as the input and outputs an embedding. The output is fed into the mask decoder together with the encoded prompt embedding. Finally, the masked decoder outputs a segmentation map which is then scaled up to the image size. The segmentation map is divided into two parts: masked region and unmasked region, which are separated by the mask boundary. The image original image is divided into: foreground object and background, which are seperated by the object boundary (shape). When the prompt is a point, the region that lies around the prompt is by default identified as the foreground, which aligns with the background. With the foreground object interpreted as the masked region, the mask prediction performance is determined by how the mask boundary aligns with the object outer shape.

\begin{figure*}[!htbp]
     \centering
         \begin{minipage}[b]{0.16\textwidth}
         \includegraphics[width=\textwidth]{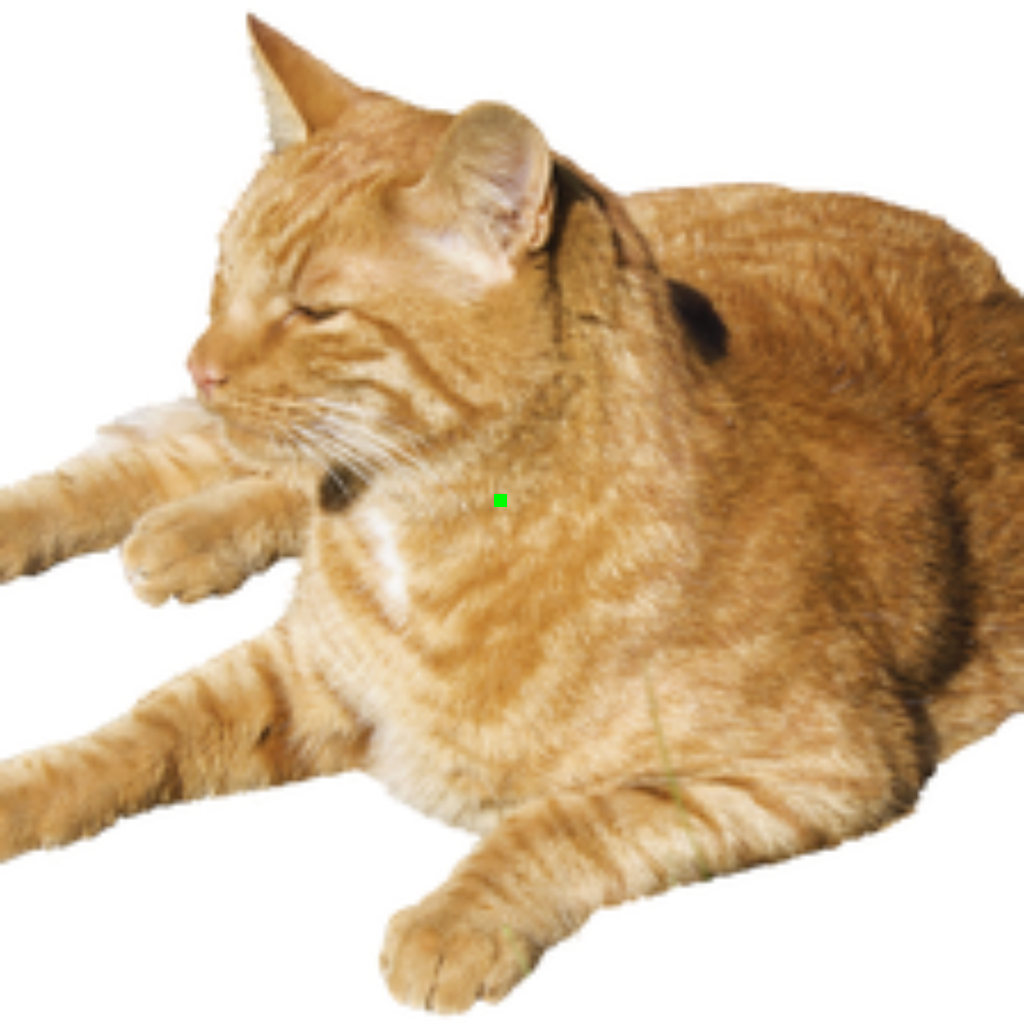}
     \end{minipage}
    \begin{minipage}[b]{0.16\textwidth}
         \includegraphics[width=\textwidth, height=1.0\textwidth]{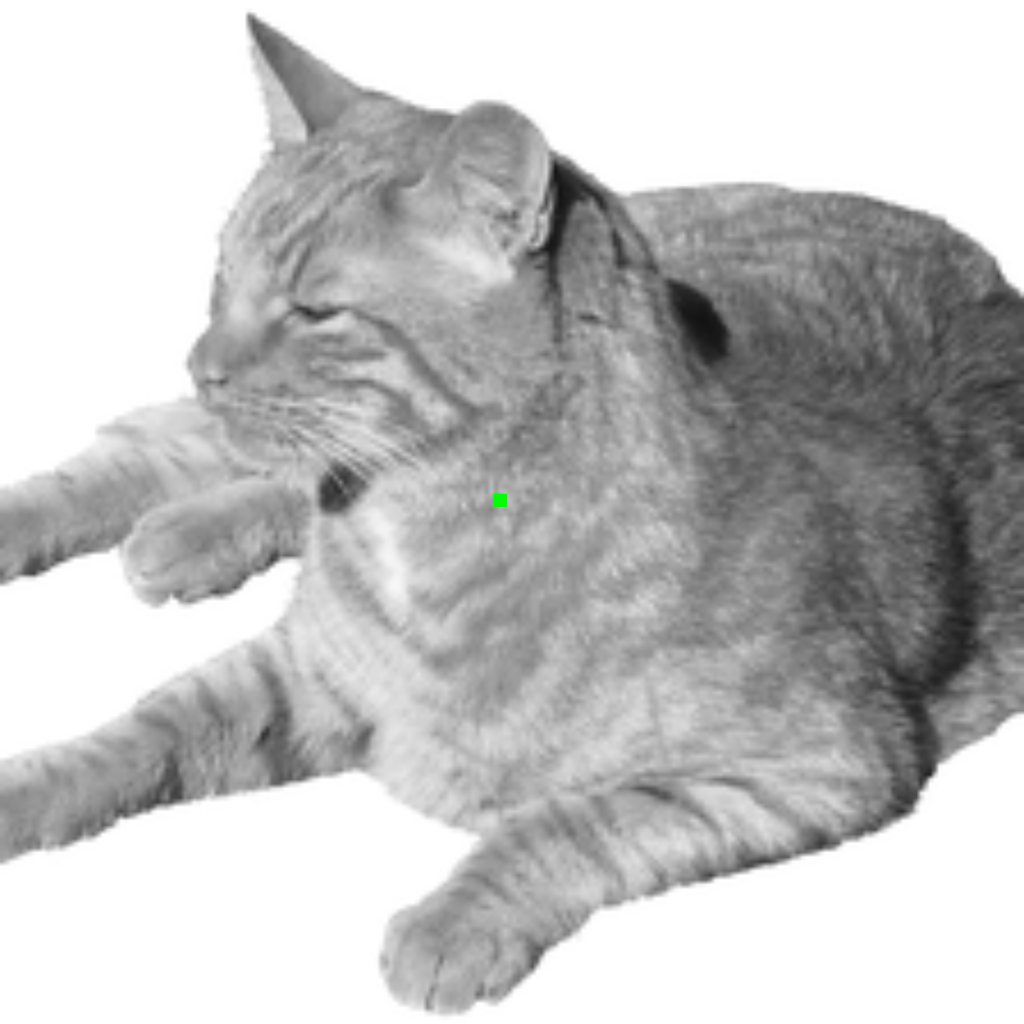}
     \end{minipage}
         \begin{minipage}[b]{0.16\textwidth}
         \centering
         \includegraphics[width=\textwidth]{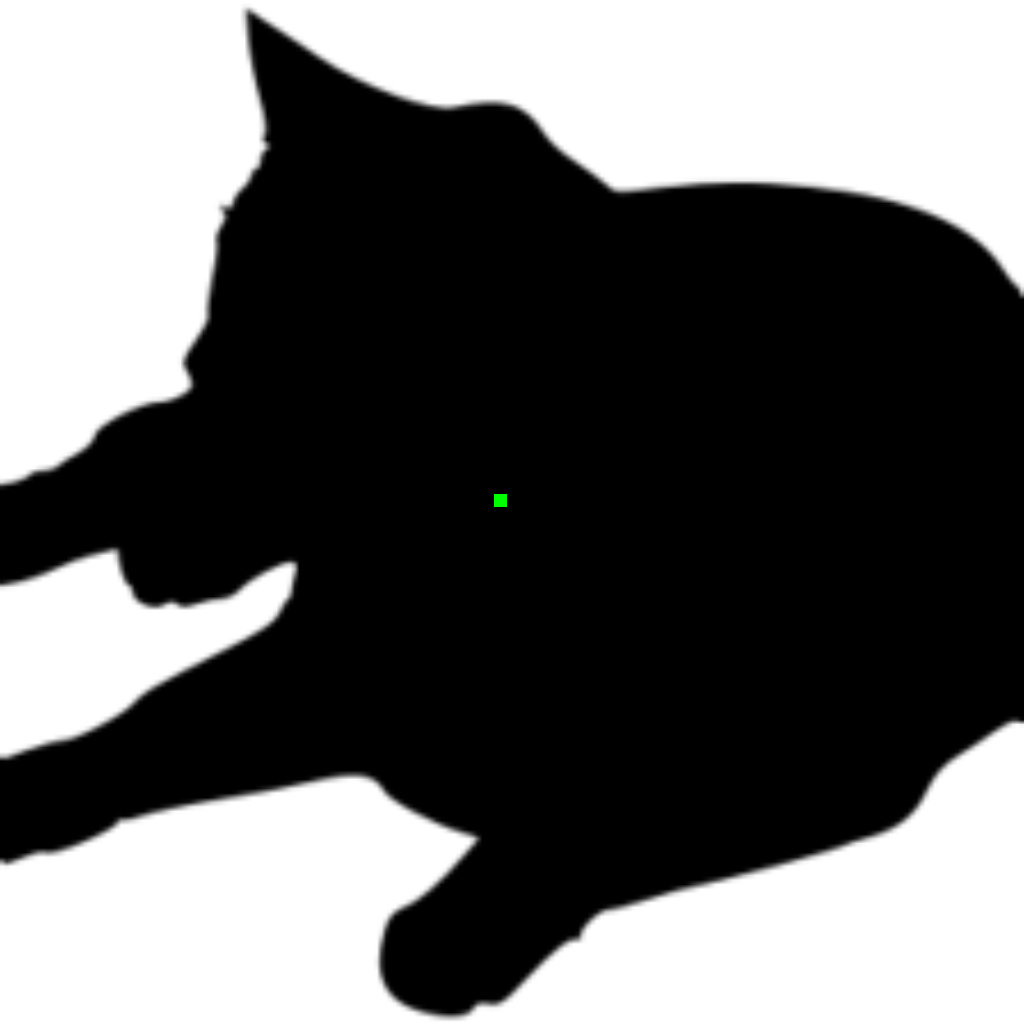}
     \end{minipage}
    \begin{minipage}[b]{0.16\textwidth}
         \centering
         \includegraphics[width=\textwidth]{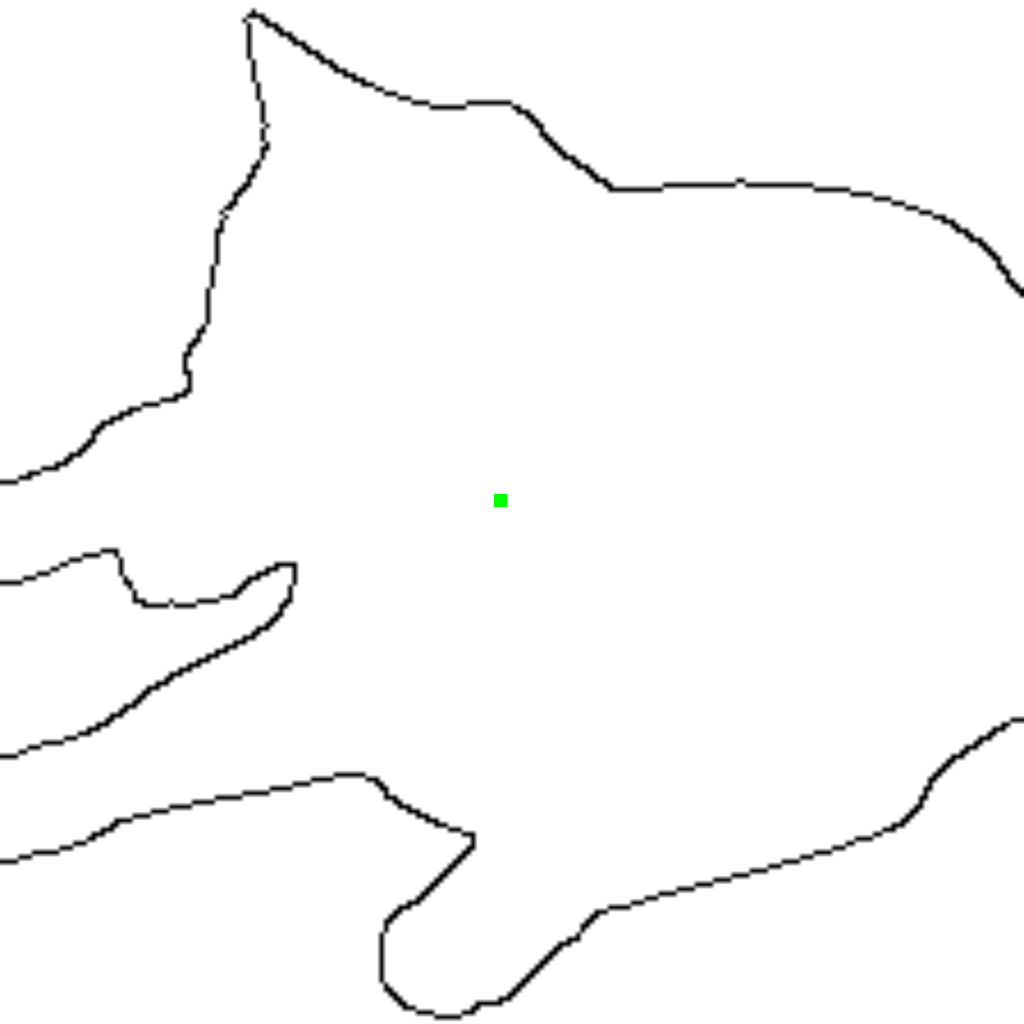}
     \end{minipage}
        \begin{minipage}[b]{0.16\textwidth}
         \centering
         \includegraphics[width=\textwidth]{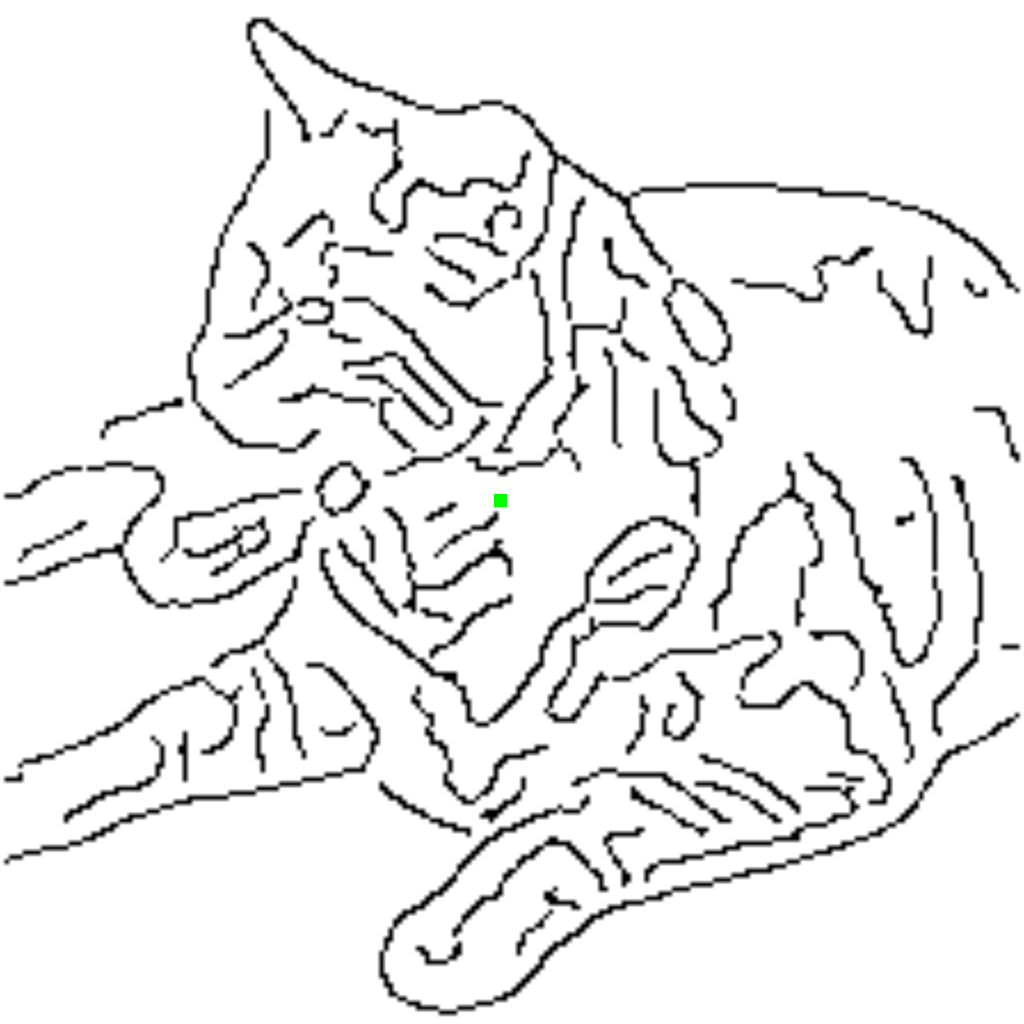}
     \end{minipage}
    \begin{minipage}[b]{0.16\textwidth}
         \centering
         \includegraphics[width=\textwidth]{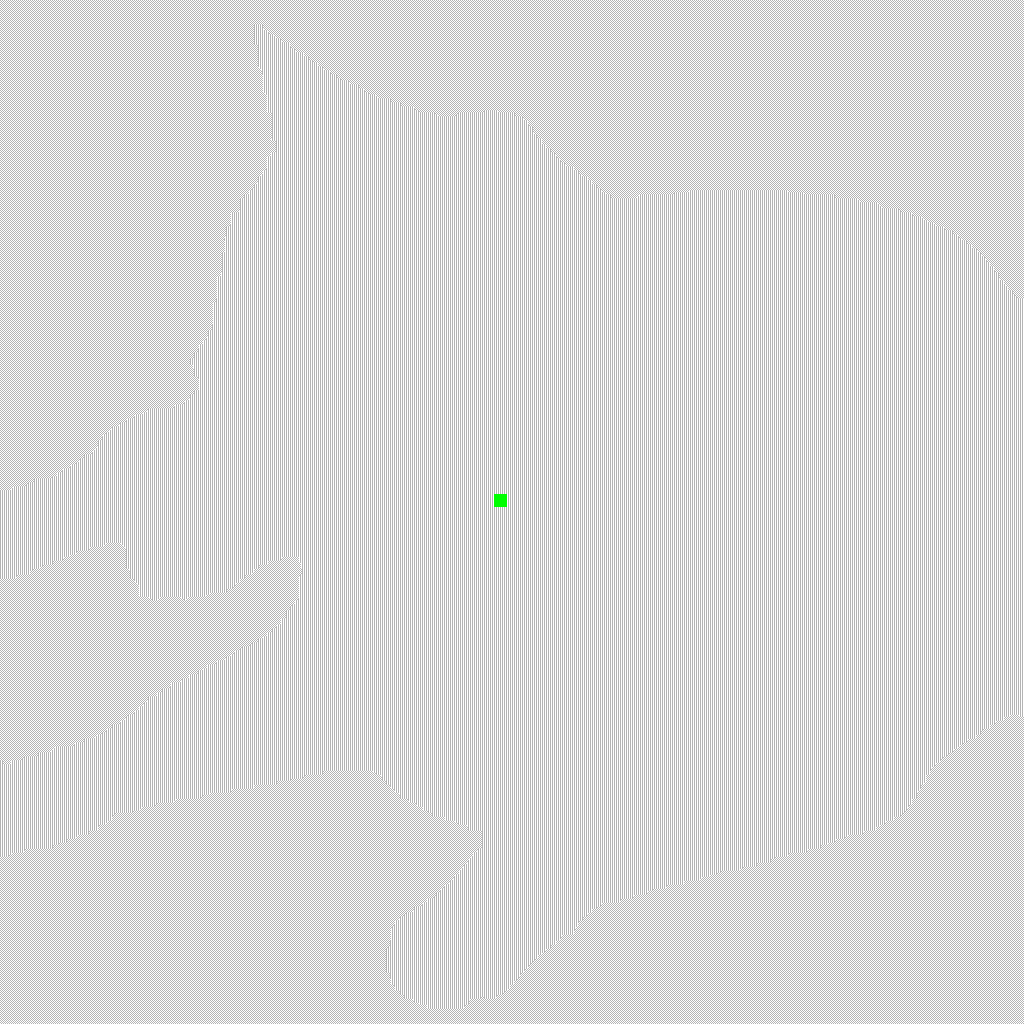}
     \end{minipage}

    \begin{minipage}[t]{0.16\textwidth}
         \includegraphics[width=\textwidth]{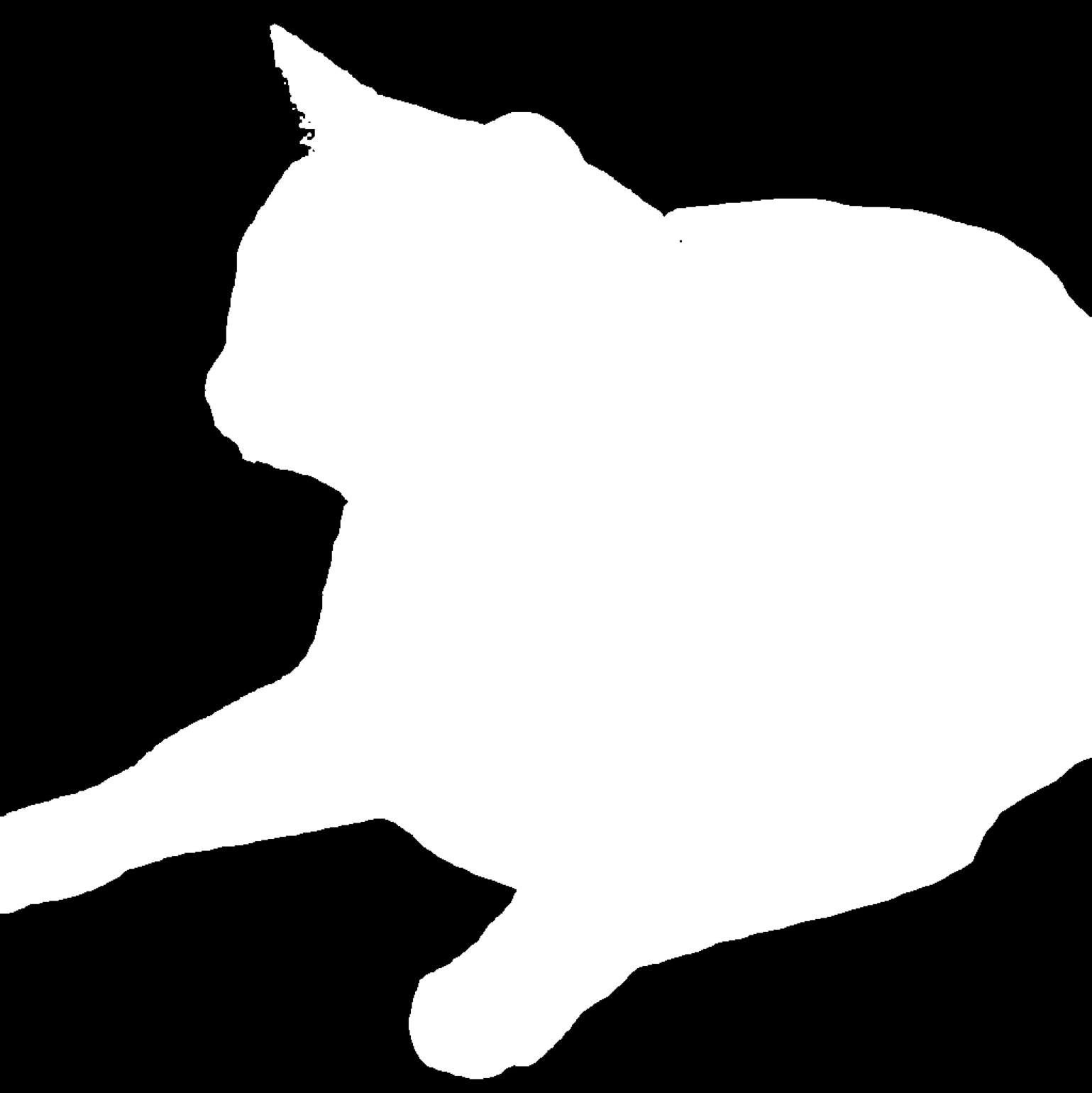}
         \subcaption{$original$}
     \end{minipage}
         \begin{minipage}[t]{0.16\textwidth}
         \centering
         \includegraphics[width=\textwidth, height=1.0\textwidth]{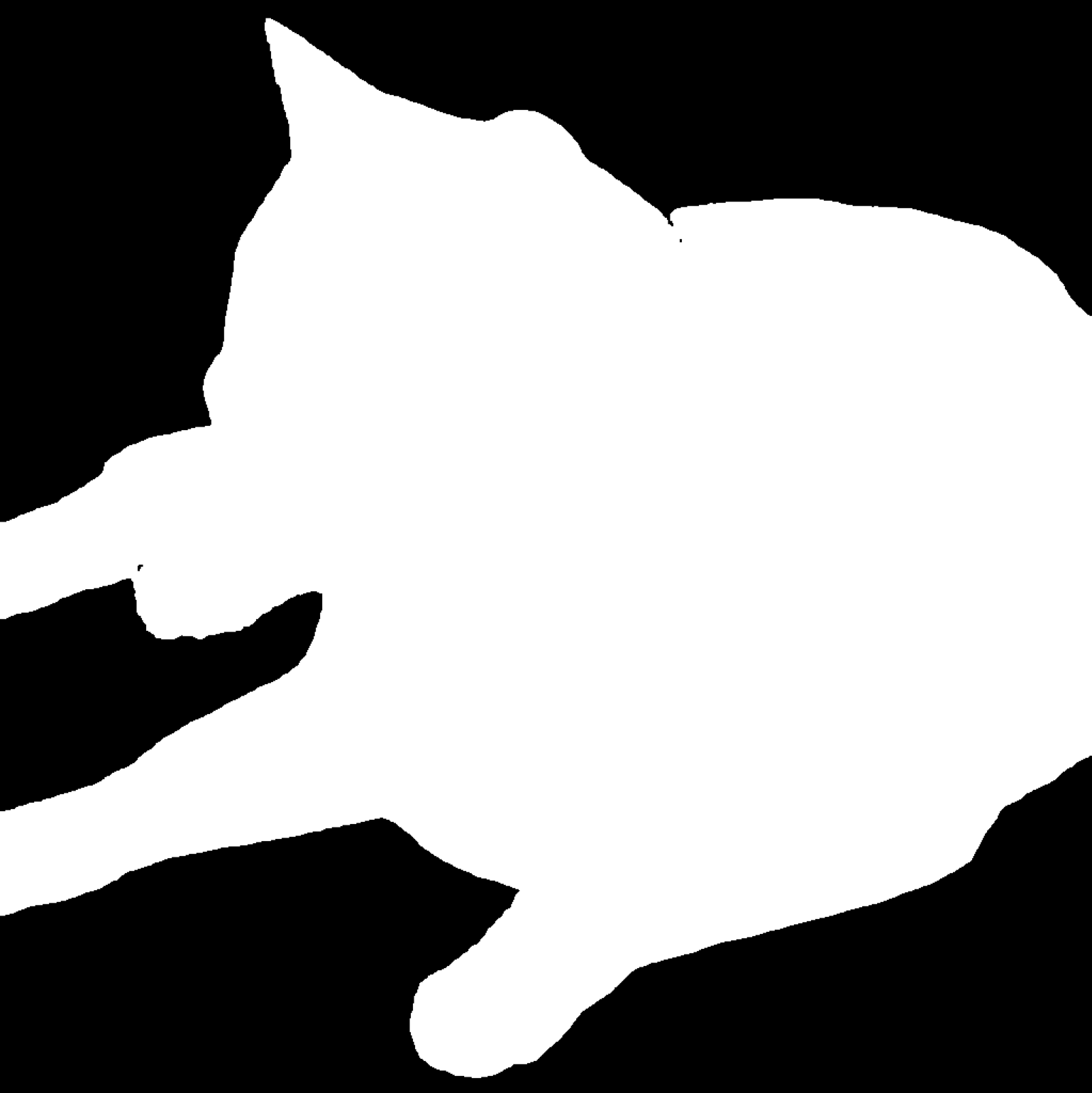}
         \subcaption{$greyscale$}
     \end{minipage}
     \begin{minipage}[t]{0.16\textwidth}
         \centering
         \includegraphics[width=\textwidth]{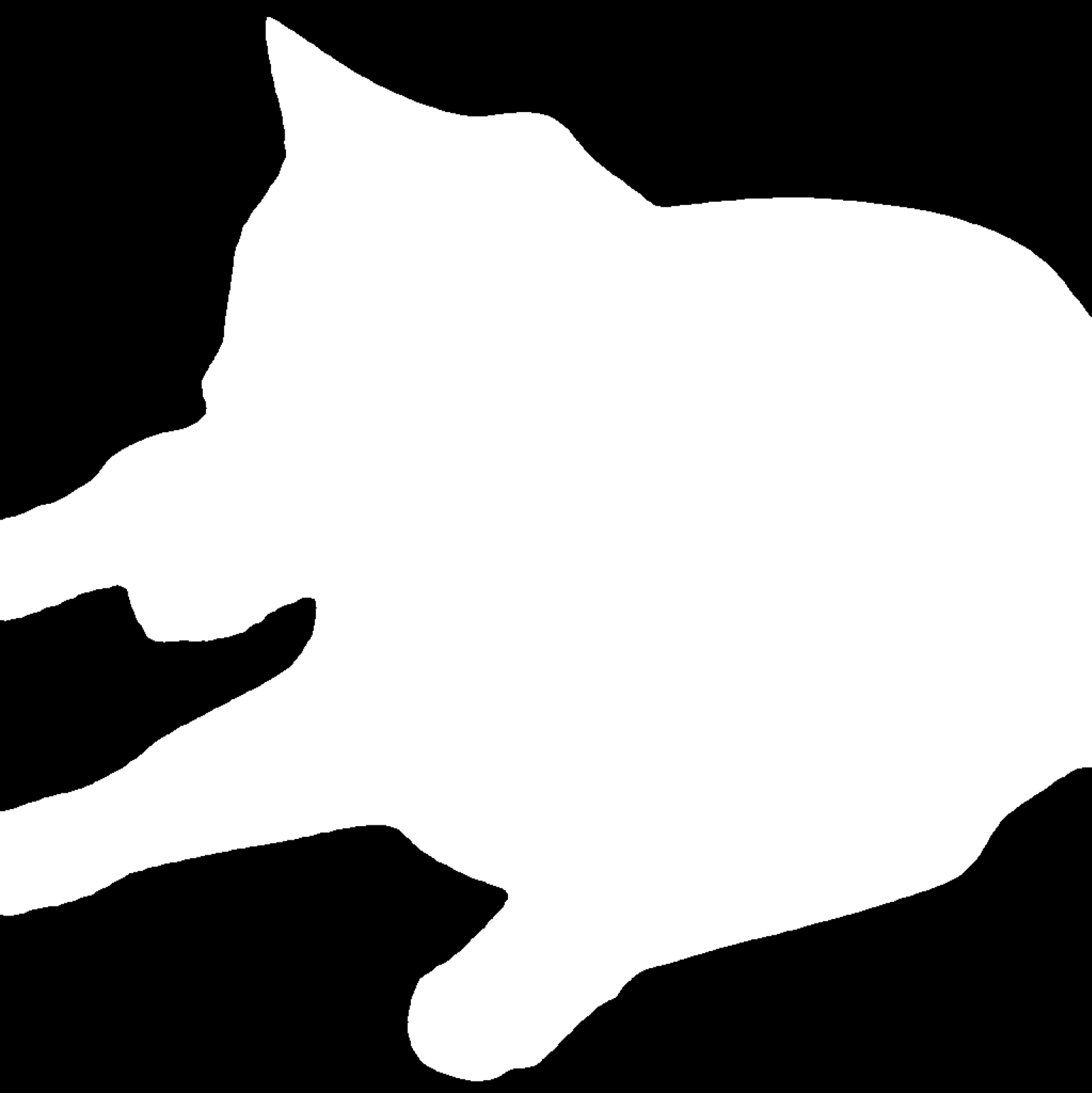}
         \subcaption{$silhouette$}
     \end{minipage}
    \begin{minipage}[t]{0.16\textwidth}
         \centering
         \includegraphics[width=\textwidth]{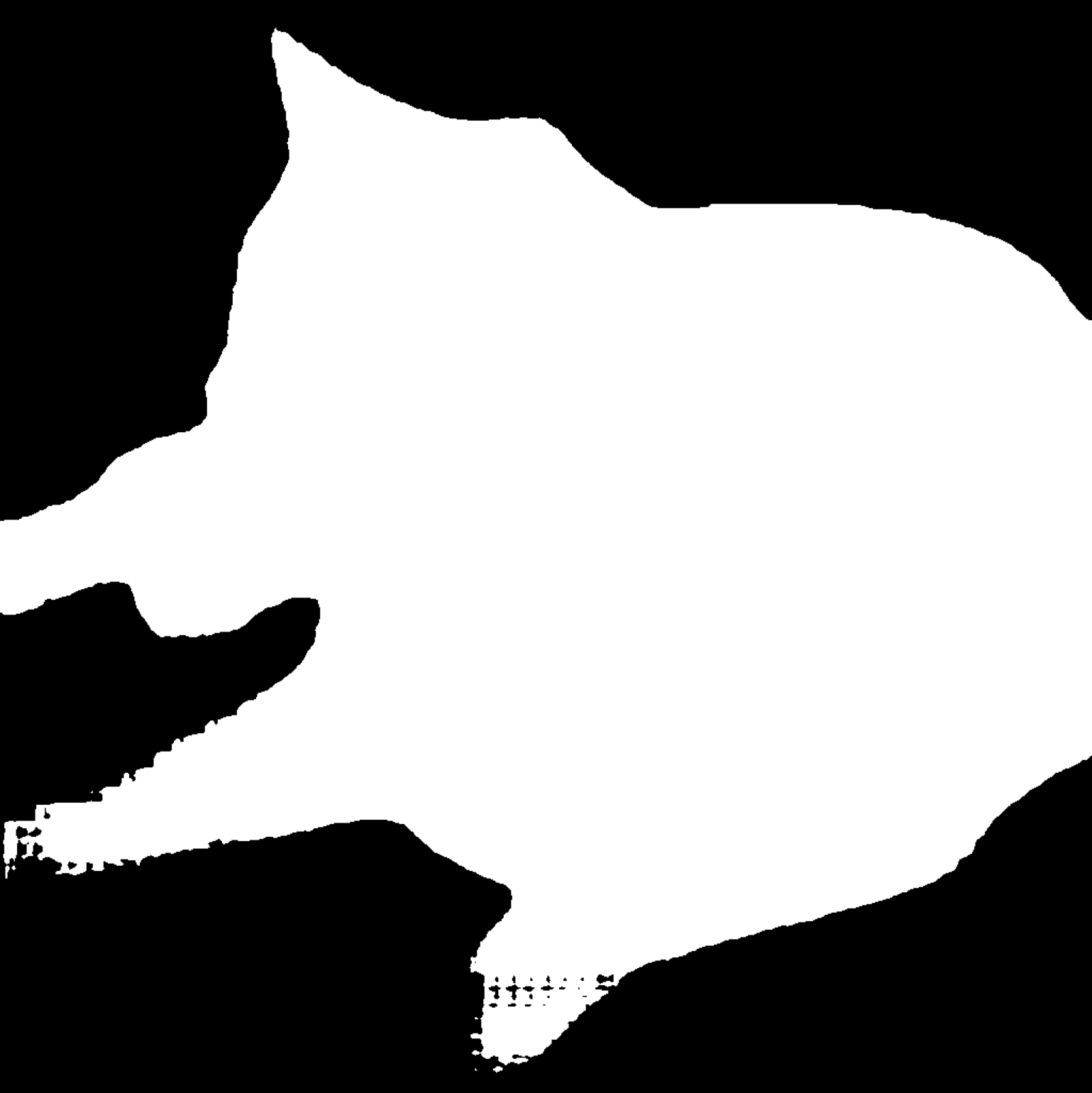}
         \subcaption{$shape$}
     \end{minipage}
        \begin{minipage}[t]{0.16\textwidth}
         \centering
         \includegraphics[width=\textwidth]{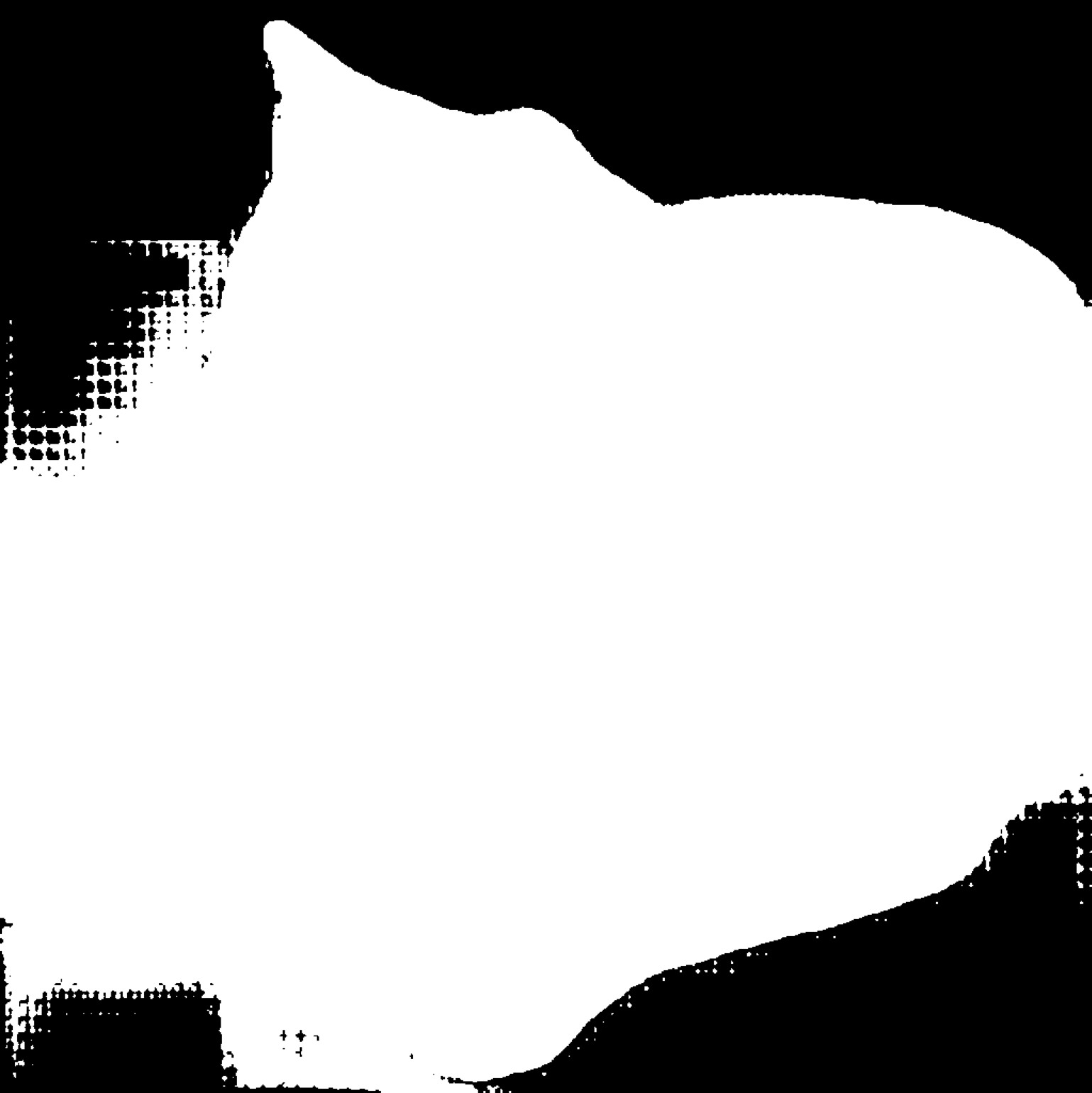}
         \subcaption{$edges$}
     \end{minipage}
     \begin{minipage}[t]{0.16\textwidth}
         \centering
         \includegraphics[width=\textwidth]{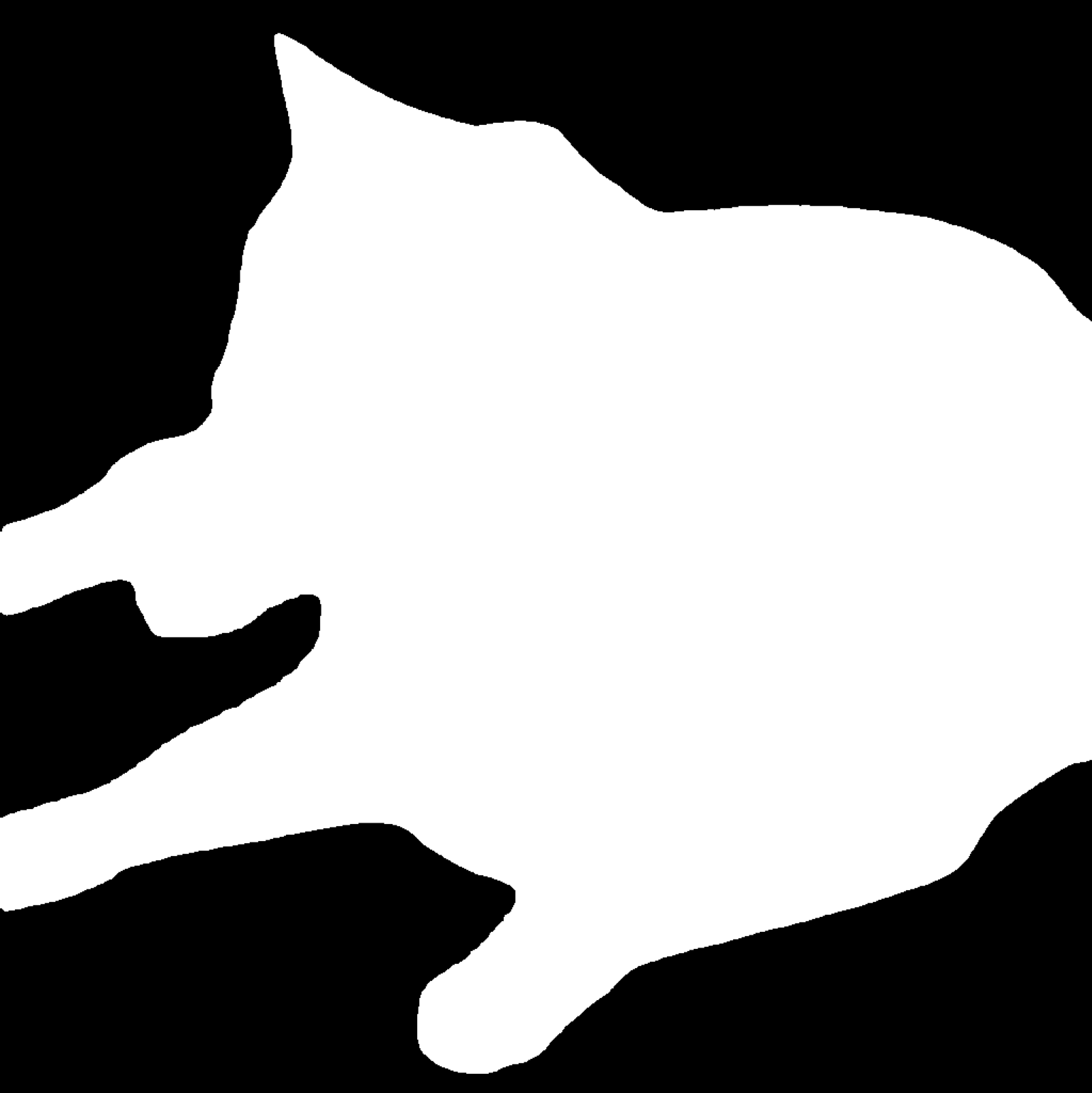}
         \subcaption{$texture$}
     \end{minipage}
    \caption{Mask prediction results of SAM under different setups. The first row indicates the images and the second row indicates the predicted mask.}
    \label{fig:cat}
\end{figure*}

In our preliminary test, we found that the original, gray scale and sihouette images give us almost the same performance of mask prediction. For simplifying the analysis, we select sihouette images as the analysis target images. A sihouette image presents the foreground as black values and the background as white values. By default, such an image provides useful cues for segmenting the object: (a) contrasting color along the object boundary (\textit{\textbf{shape}} ); (b) different \textbf{\textit{texture}} content in the foreground and background.


\begin{figure*}[!htbp]
     \centering
         \begin{minipage}[b]{0.20\textwidth}
         \includegraphics[width=\textwidth]{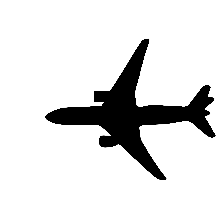}
     \end{minipage}
    \begin{minipage}[b]{0.20\textwidth}
         \includegraphics[width=\textwidth, height=1.0\textwidth]{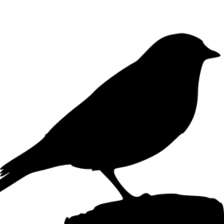}
     \end{minipage}
         \begin{minipage}[b]{0.20\textwidth}
         \centering
         \includegraphics[width=\textwidth]{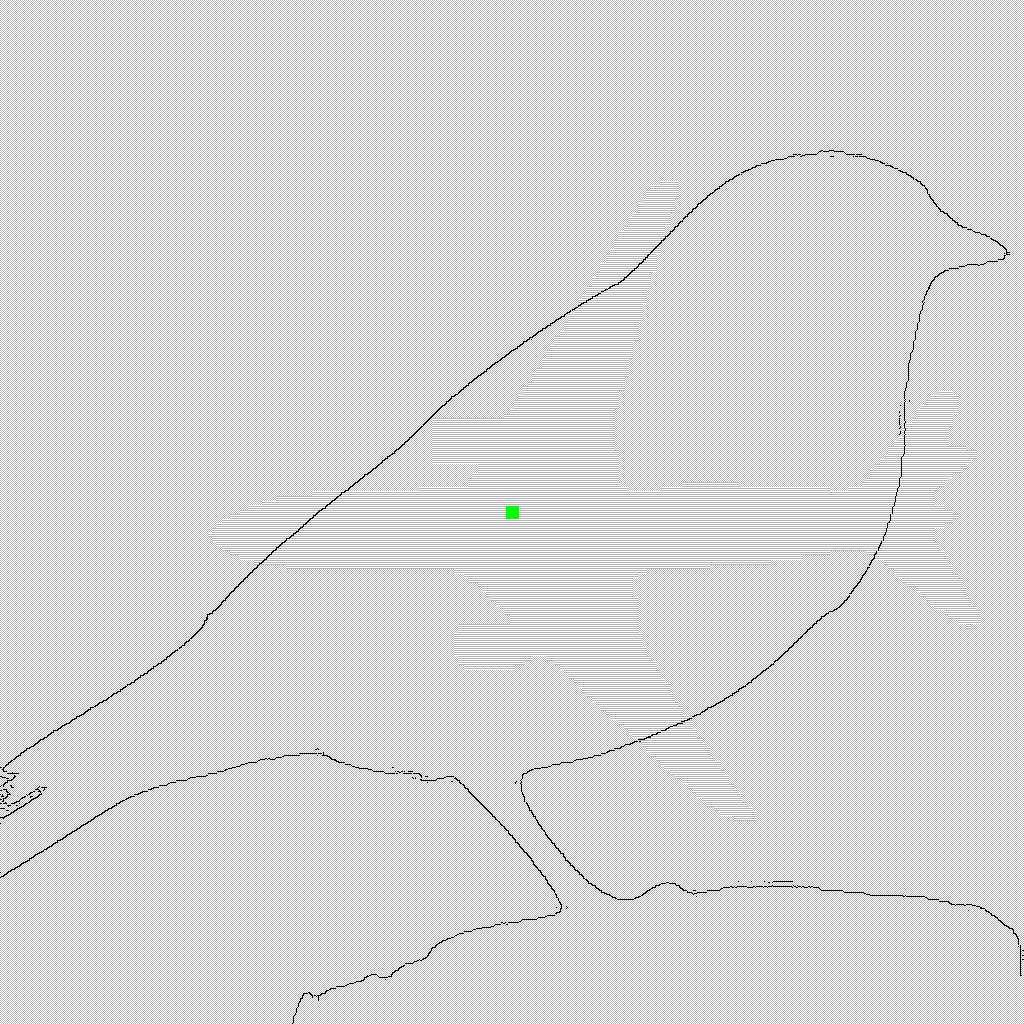}
     \end{minipage}
        \begin{minipage}[b]{0.20\textwidth}
         \centering
         \includegraphics[width=\textwidth]{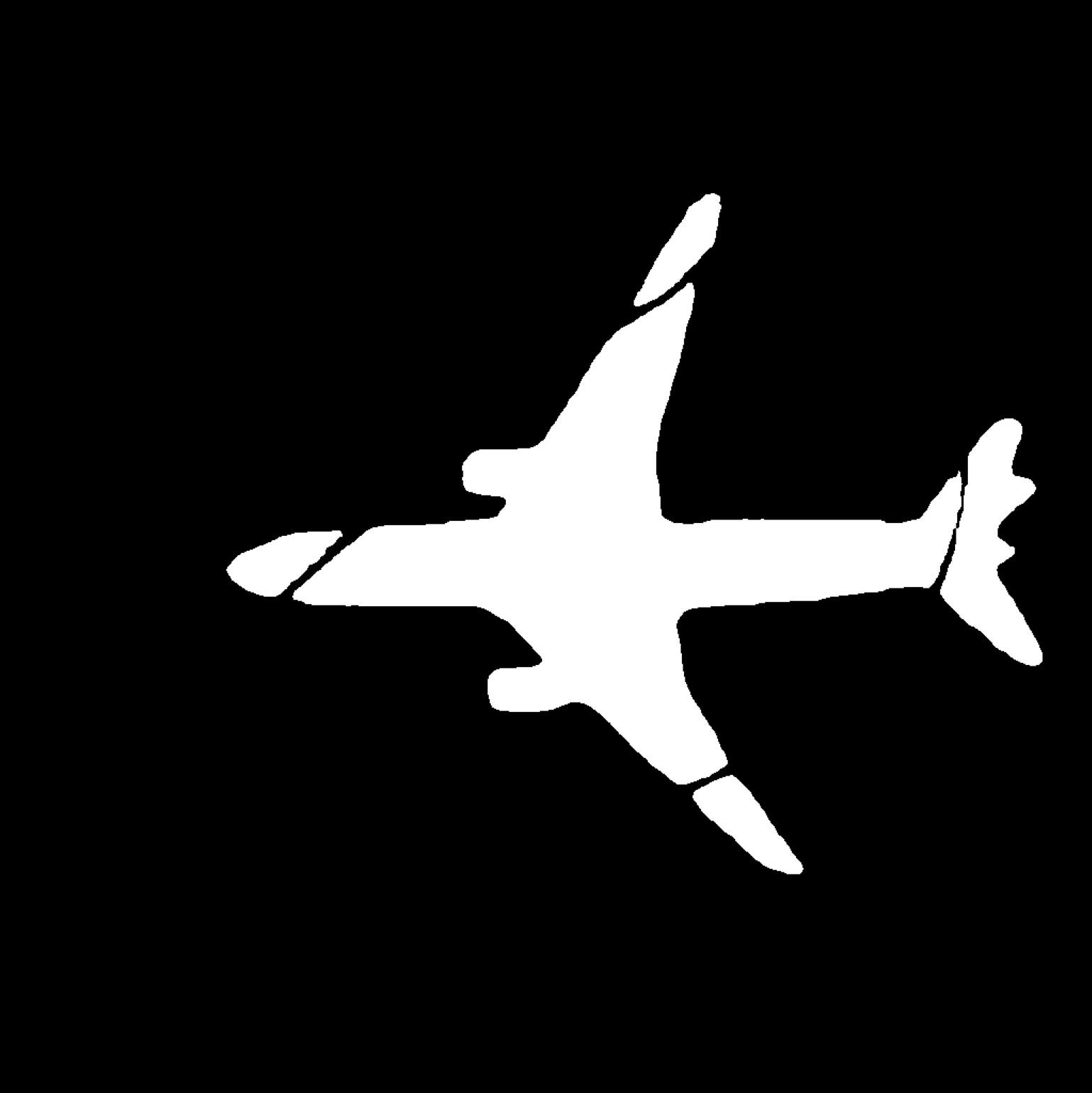}
     \end{minipage}

    \begin{minipage}[b]{0.20\textwidth}
         \includegraphics[width=\textwidth]{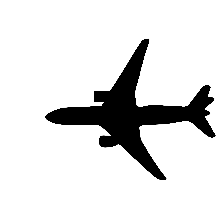}
     \end{minipage}
    \begin{minipage}[b]{0.20\textwidth}
         \includegraphics[width=\textwidth, height=1.0\textwidth]{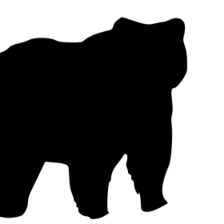}
     \end{minipage}
         \begin{minipage}[b]{0.20\textwidth}
         \centering
         \includegraphics[width=\textwidth]{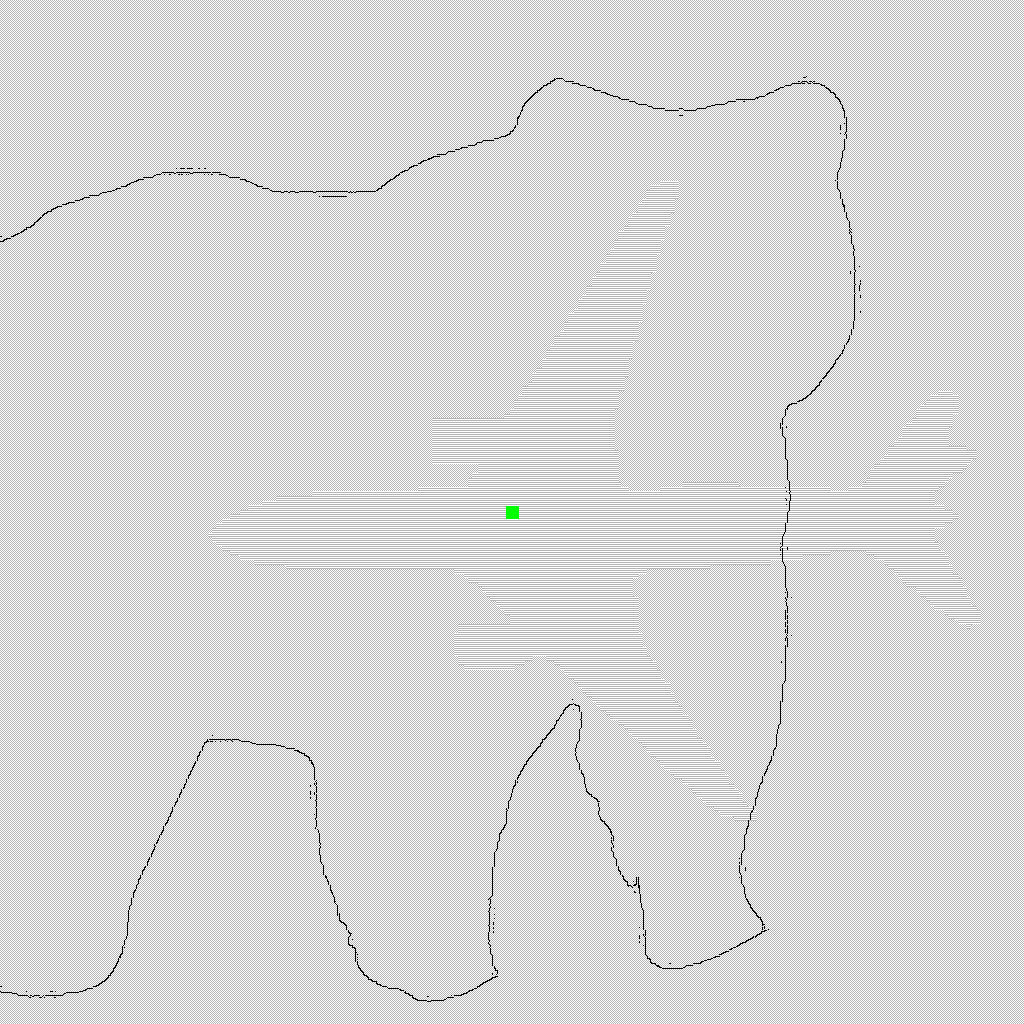}
     \end{minipage}
        \begin{minipage}[b]{0.20\textwidth}
         \centering
         \includegraphics[width=\textwidth]{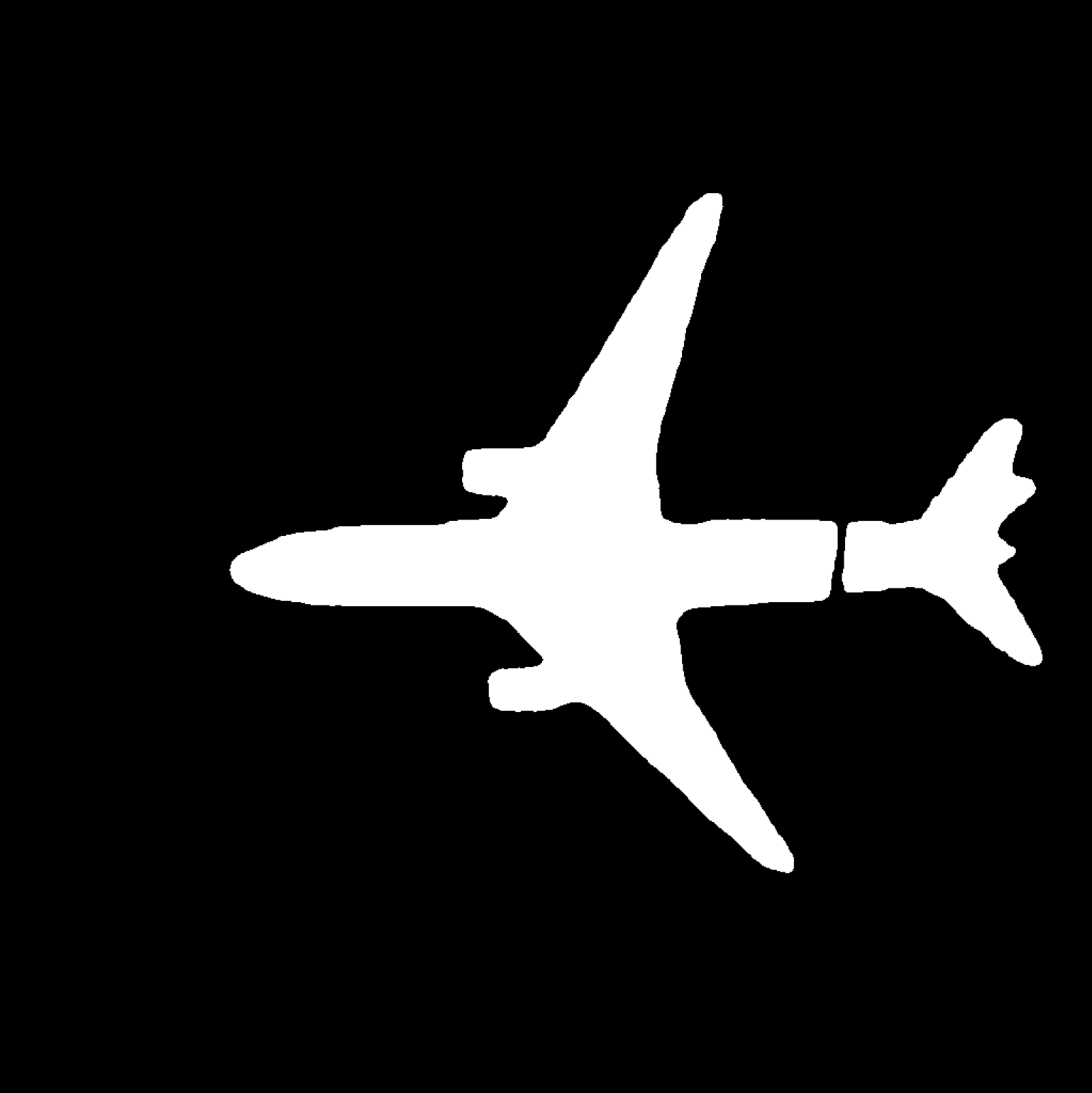}
     \end{minipage}

    \begin{minipage}[b]{0.20\textwidth}
         \includegraphics[width=\textwidth, height=1.0\textwidth]{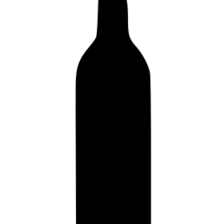}
     \end{minipage}
        \begin{minipage}[b]{0.20\textwidth}
         \includegraphics[width=\textwidth]{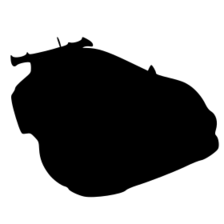}
     \end{minipage}
         \begin{minipage}[b]{0.20\textwidth}
         \centering
         \includegraphics[width=\textwidth]{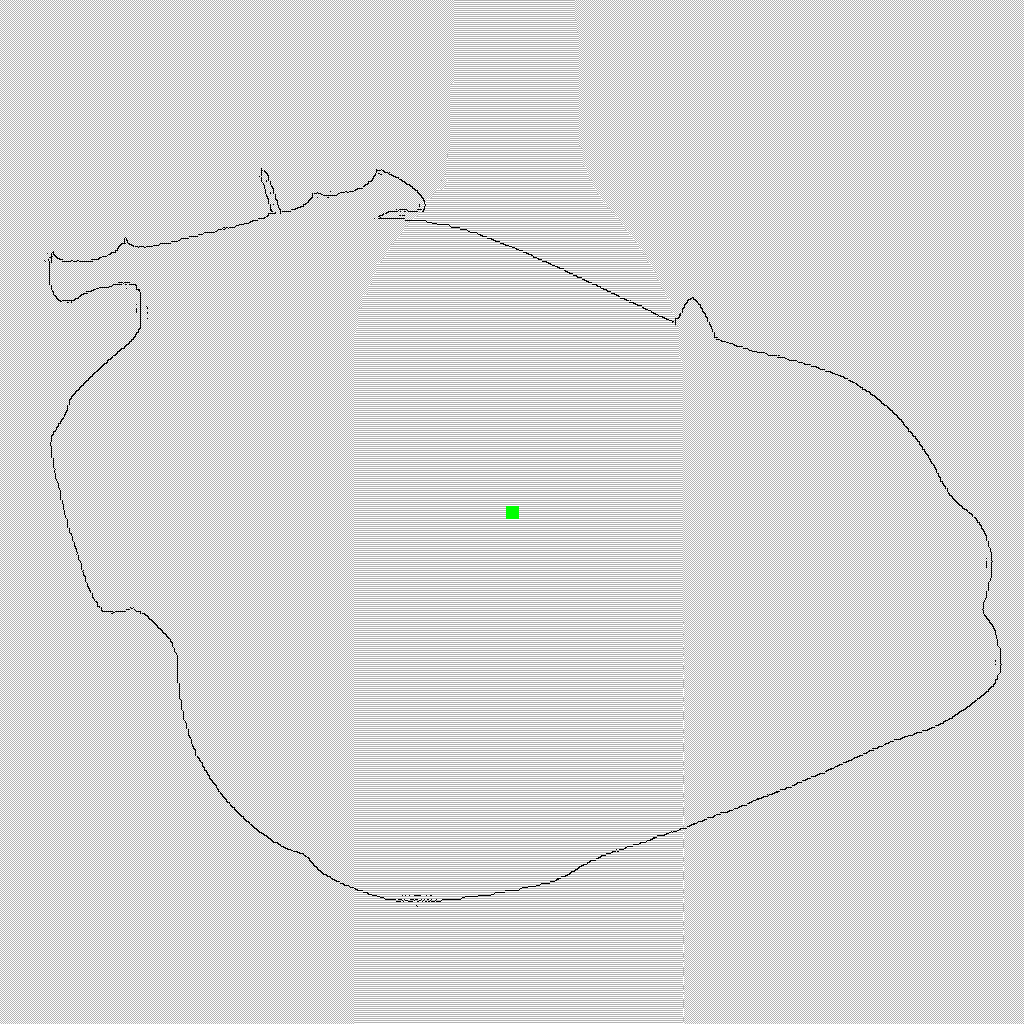}
     \end{minipage}
        \begin{minipage}[b]{0.20\textwidth}
         \centering
         \includegraphics[width=\textwidth]{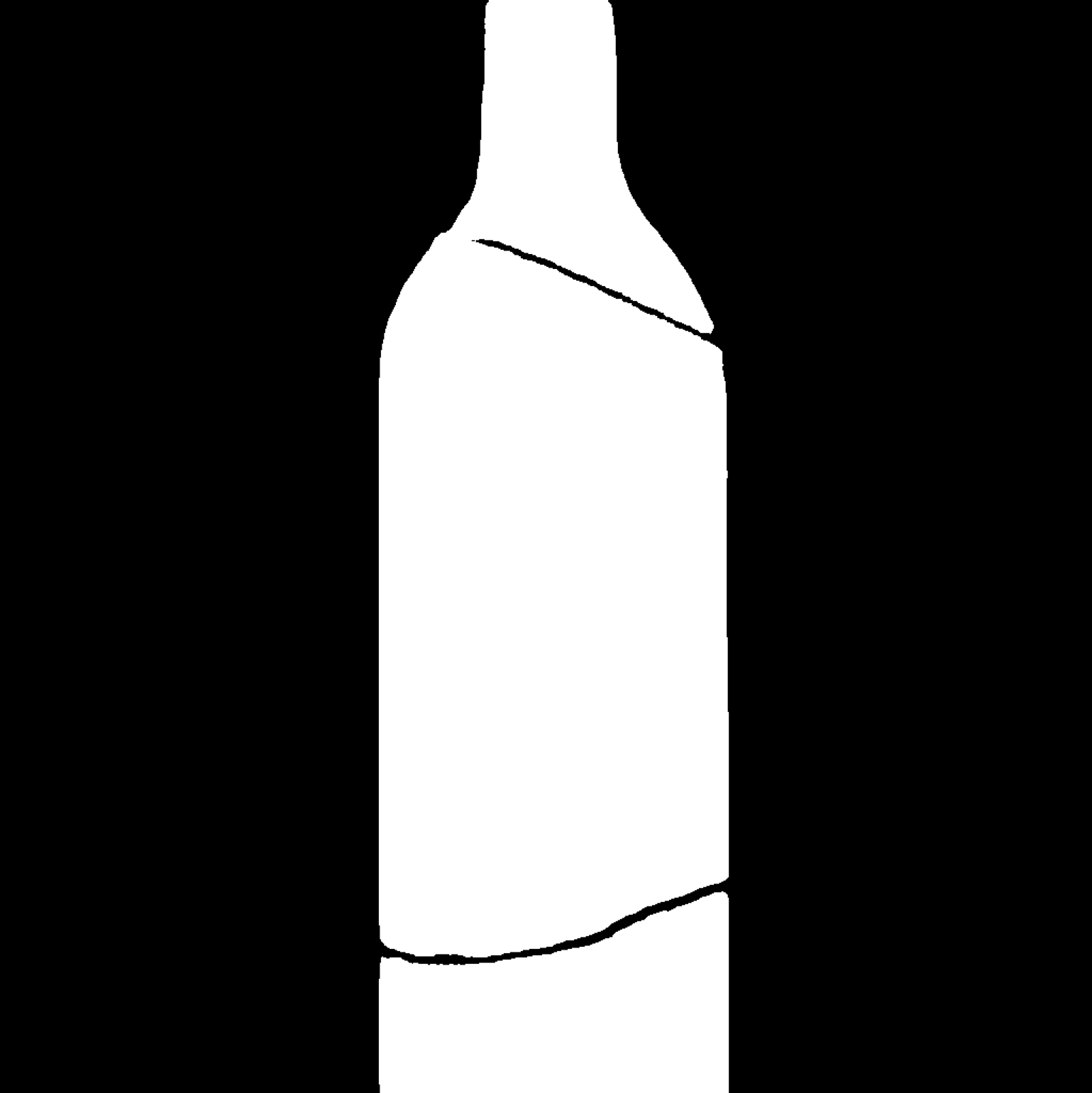}
     \end{minipage}

    \begin{minipage}[b]{0.20\textwidth}
         \includegraphics[width=\textwidth]{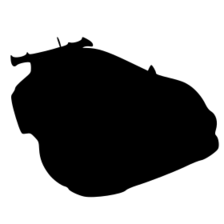}
     \end{minipage}
    \begin{minipage}[b]{0.20\textwidth}
         \includegraphics[width=\textwidth, height=1.0\textwidth]{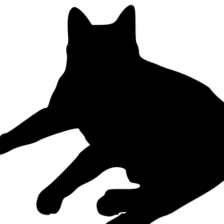}
     \end{minipage}
         \begin{minipage}[b]{0.20\textwidth}
         \centering
         \includegraphics[width=\textwidth]{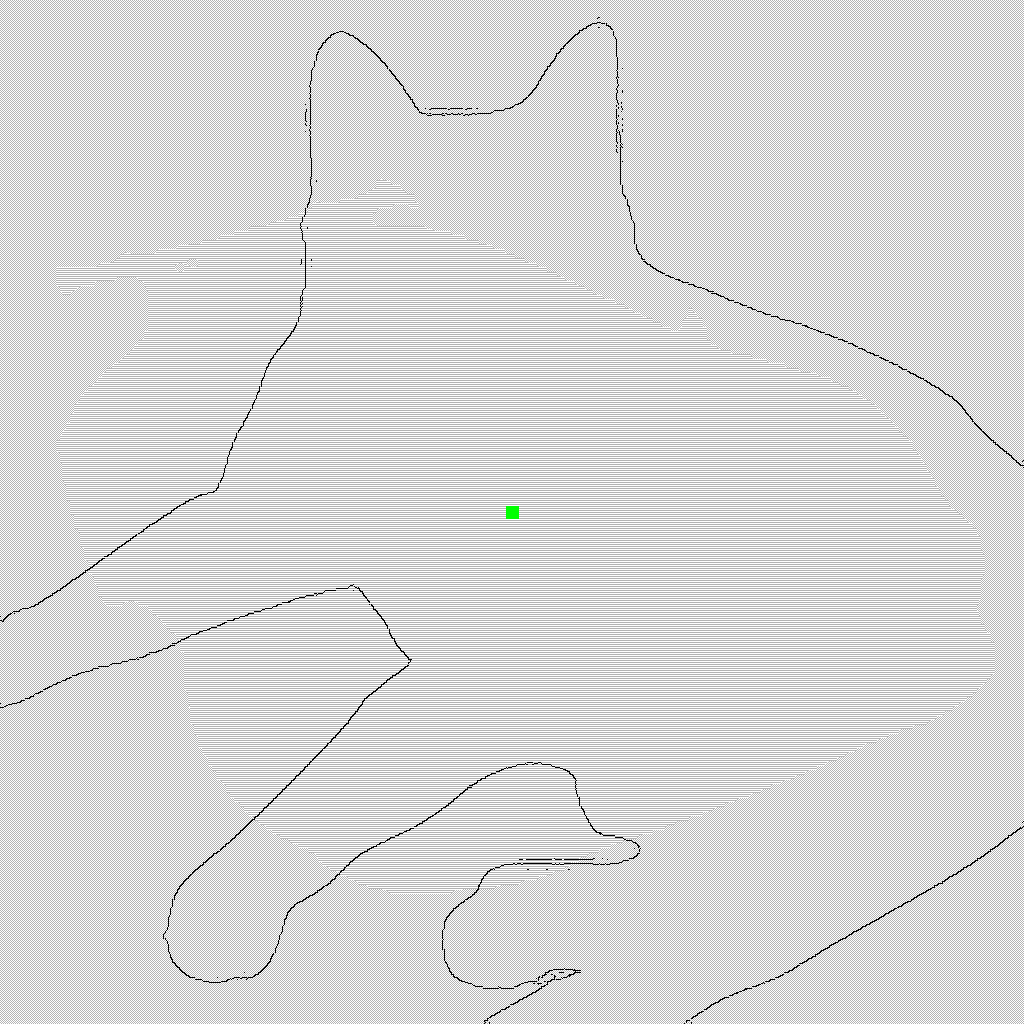}
     \end{minipage}
        \begin{minipage}[b]{0.20\textwidth}
         \centering
         \includegraphics[width=\textwidth]{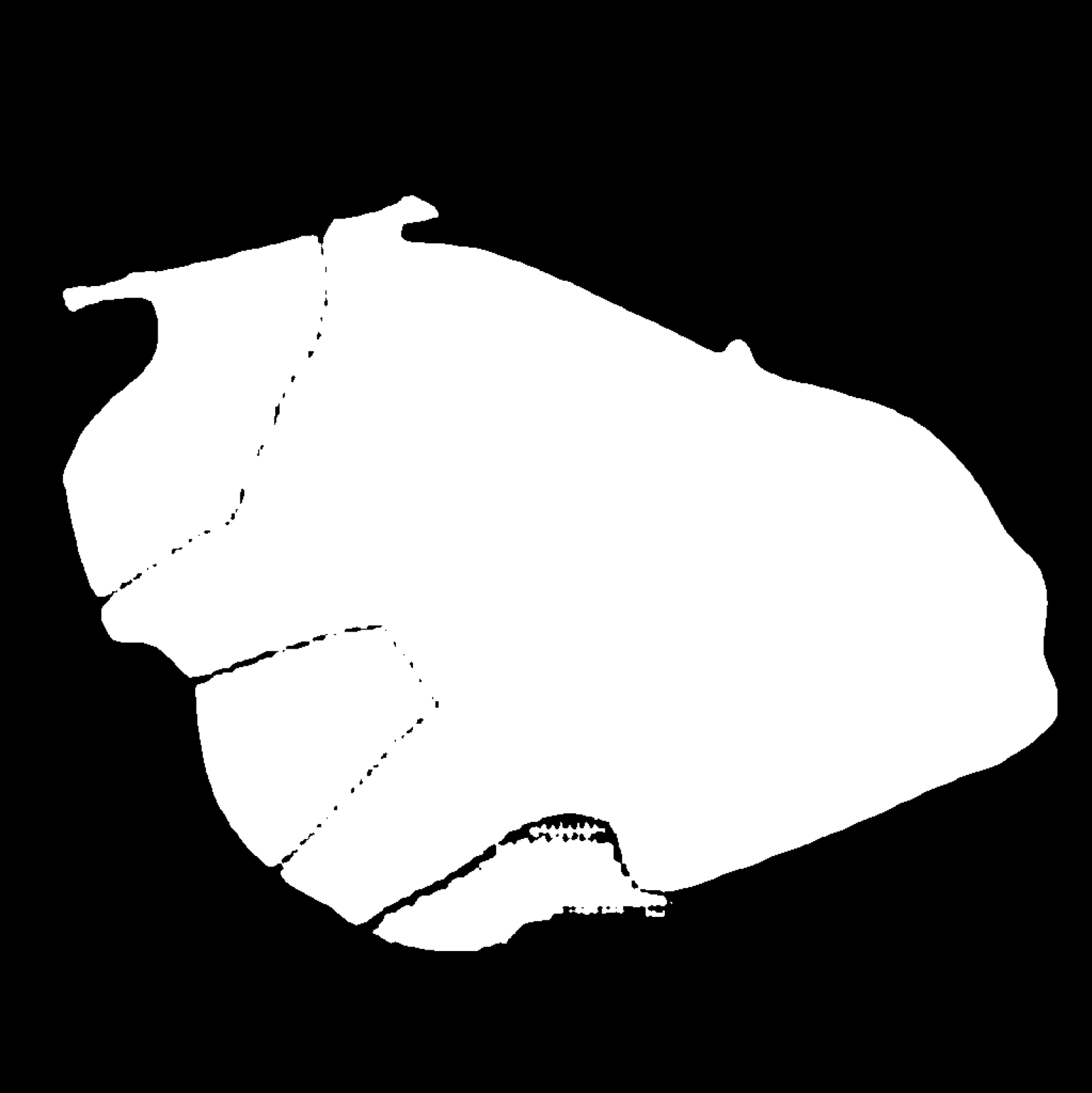}
     \end{minipage}

    \begin{minipage}[b]{0.20\textwidth}
         \includegraphics[width=\textwidth]{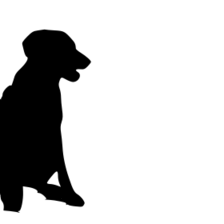}
     \end{minipage}
    \begin{minipage}[b]{0.20\textwidth}
         \includegraphics[width=\textwidth, height=1.0\textwidth]{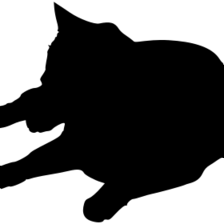}
     \end{minipage}
         \begin{minipage}[b]{0.20\textwidth}
         \centering
         \includegraphics[width=\textwidth]{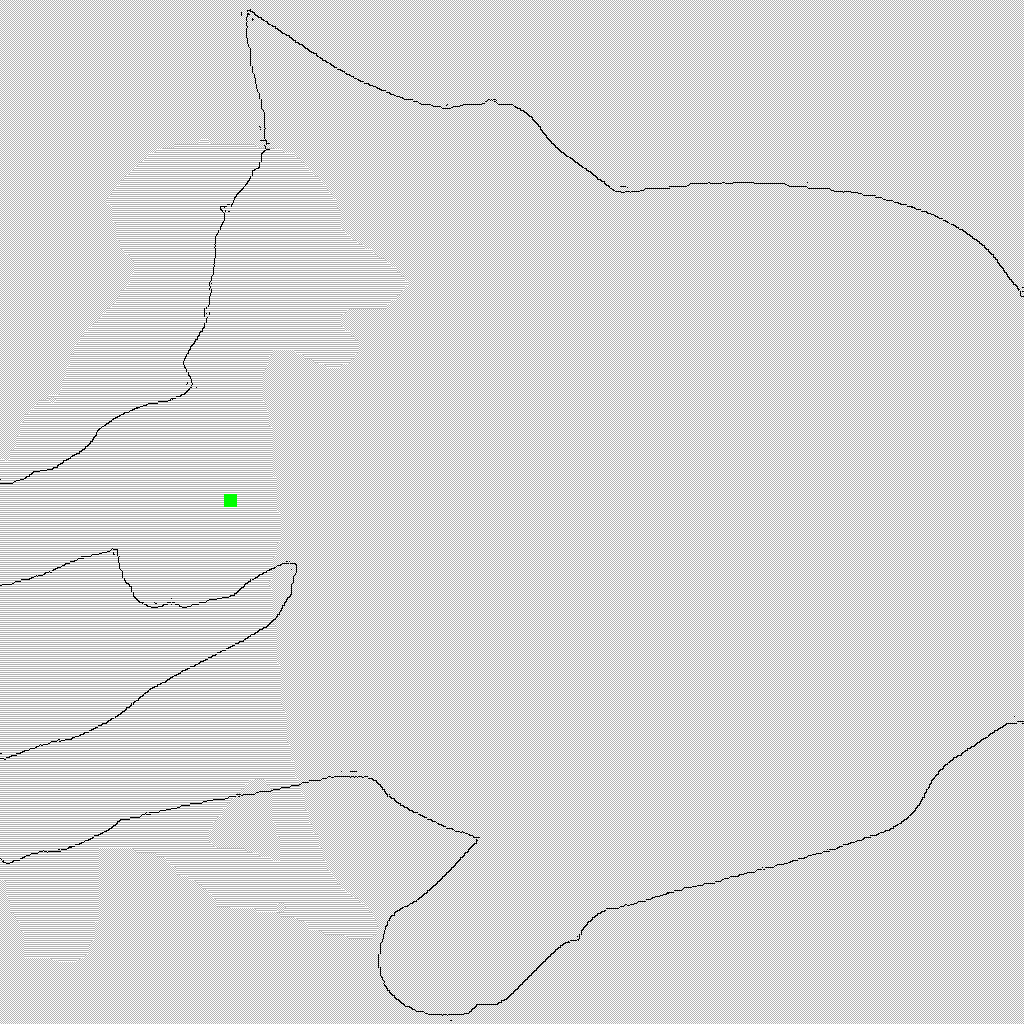}
     \end{minipage}
        \begin{minipage}[b]{0.20\textwidth}
         \centering
         \includegraphics[width=\textwidth]{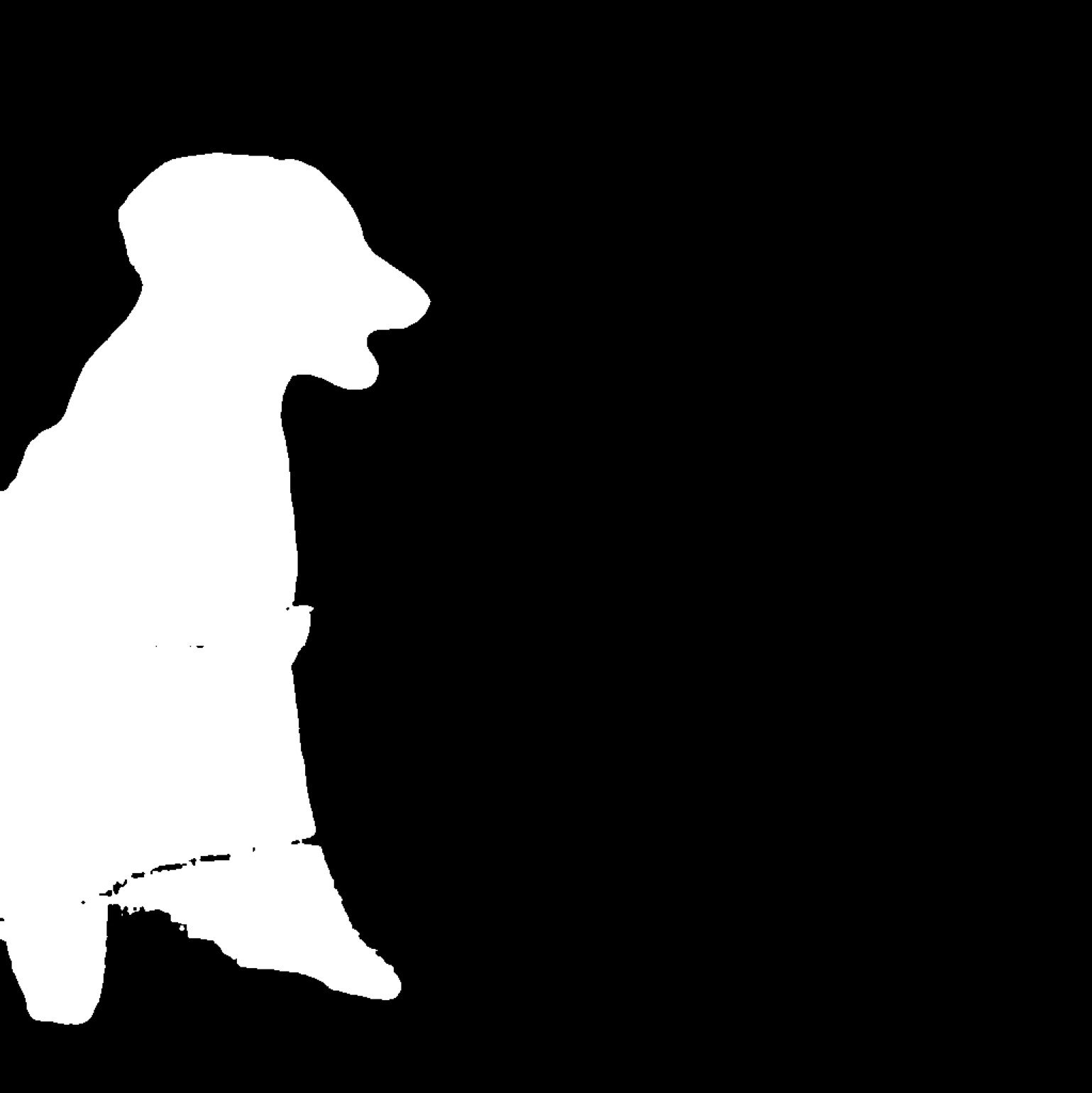}
     \end{minipage}

     \begin{minipage}[b]{0.20\textwidth}
         \includegraphics[width=\textwidth]{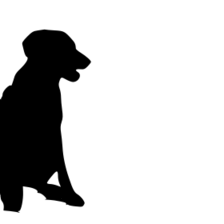}
     \end{minipage}
    \begin{minipage}[b]{0.20\textwidth}
         \includegraphics[width=\textwidth, height=1.0\textwidth]{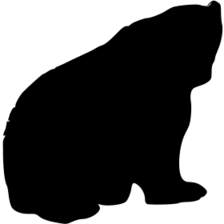}
     \end{minipage}
         \begin{minipage}[b]{0.20\textwidth}
         \centering
         \includegraphics[width=\textwidth]{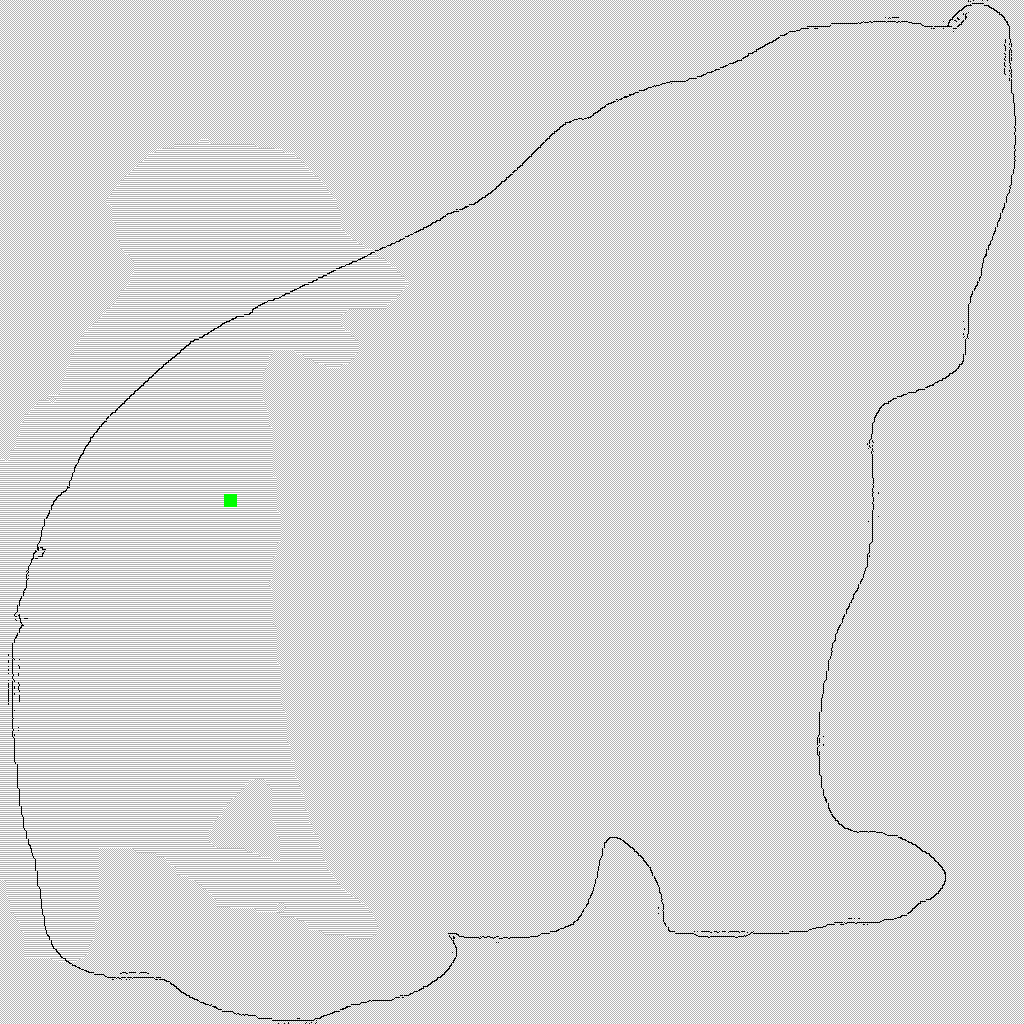}
     \end{minipage}
        \begin{minipage}[b]{0.20\textwidth}
         \centering
         \includegraphics[width=\textwidth]{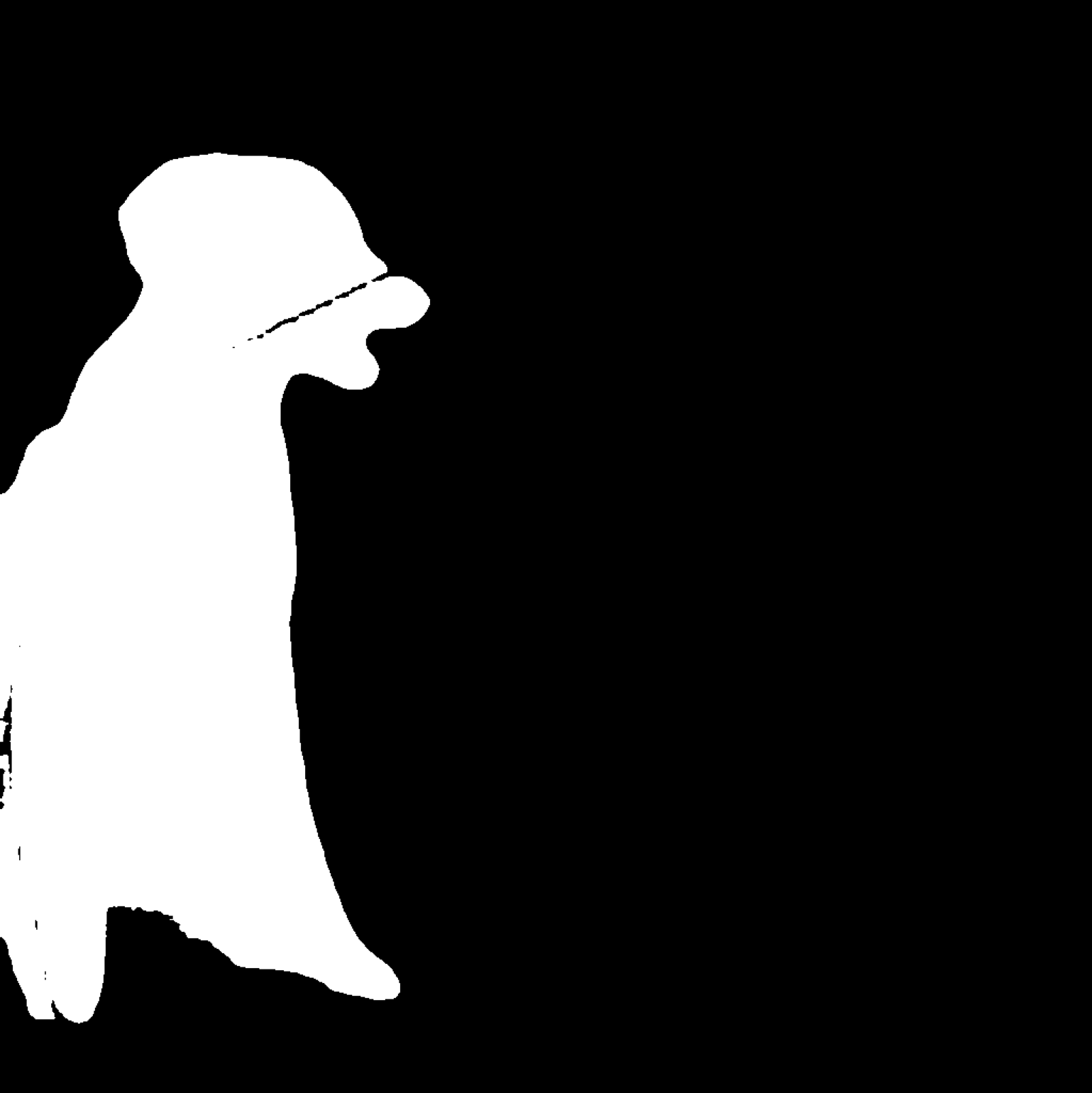}
     \end{minipage}

        \caption{Mask prediction results of SAM under shape-texture conflict. The leftmost two columns show the images utilized for generating texture and shapes, respectively. The third column indicates the image with the designed shape-texture conflict. The last column shows the predicted mask.}
    \label{fig:conflict_results}
\end{figure*}

\textbf{How to design texture-shape cue conflict?} To determine which cue plays a more dominant role for deciding the mask prediction, we need to design a texture-shape cue conflict. Prior work~\cite{geirhos2018imagenet} designs a texture-shape cue conflict for label prediction by blending shape of one image with texture of the other. Such an approach, however, cannot generate the desired texture-shape cue conflict for recognizing the object mask because the blended image does not have texture cue. To create texture-shape cue conflict for mask prediction, we first design patterns that either have shape cue or texture cue. The shape-only cue can be simply obtained by keeping the boundary while making the foreground have the same texture content as the background. To obtain texture-only cue without creating an obvious color contrast (along the object boundary shape), we design the texture pattern with stripe-like content or checkboard. Since the texture pattern alternates with different colors in high frequency, the two texture types do not have a clear color contrast along their boundary. To make foreground and background have different textures, we can make them have different styles (like one using stripes and the other using checkboard). In the following, we elaborate on how to create texture-shape cue conflict for mask prediction with a concrete example.

\textbf{A concrete example: texture cue for bear and shape cue for cat.} We first generate a texture cue for bear by selecting a bear image. We fill the foreground area with texture of one style and fill the background area with texture of another style. As we can see from Figure~\ref{fig:conflict_setup}, there is no clear color contrast between the foreground and background, thus avoiding the shape cue. On top of this, we further blend it with a shape-only cue for cat.

\section{Results}
\textbf{Non-conflict cues.} Here, we take an image of cat used in~\cite{geirhos2018imagenet} for analysis and the results are shown in Figure~\ref{fig:cat}. We find that the SAM predicts the mask of the cat object well, which is well expected. Similar performance can be observed for greyscale image because it still contains both shape and texture for predicting masks. When the texture is removed as in the shape image, the quality of the predicted mask decreases. This phenomenon gets more pronounced for edge images, which resembles the finding in~\cite{geirhos2018imagenet} that deep classification models recognize the shape images better than corresponding edge images. The quality of the predicted mask with texture cue is very close to that of original image. This result suggests that texture-only cue might be sufficient for mask prediction when there is no clear shape-cue along the object boundary. For completeness, we also report the result of silhouette image, which is well expected since it contains both shape and texture cues.

\textbf{Texture-shape cue conflict.} The results of texture-shape cue conflict experiments are shown in Figure~\ref{fig:conflict_results}. We find that the mask predicted by the SAM is dominated by the texture cue rather than shape cue in most cases. For example, in the  first row of Figure~\ref{fig:conflict_results}, the human can easily detect a bird in the blended image due to the bird shape, however, the SAM predicts the airplane as the mask due to its texture cue. It is worth mentioning that the predicted mask is still influenced by the bird shape to some extent especially in those regions where the mask region overlaps with the bird shape. Other rows of example cases show a similar trend. Overall, the results in Figure~\ref{fig:conflict_results} support that the SAM is biased towards texture bias rather than shape.

\section{Conclusion}
In this work, we are the first to understand segment anything model (SAM). Specifically, we explain how SAM achieves impressive mask prediction by disentangling texture and shape cues. We find texture-only cue is often sufficient for well predicting the mask, while the shape-only cue is less predictive. Moreover, we demonstrate that in most texture-shape cue conflict setups, the SAM prediction is often dominated by the texture cue rather than shape cue. 

\bibliographystyle{unsrtnat}
\bibliography{bib_mixed,bib_local,bib_sam}

\end{document}